\pdfoutput=1

\documentclass[11pt]{article}
\makeatletter
\renewcommand{\@makefntext}[1]{%
  \noindent
  \makebox[5pt][r]{\@makefnmark}#1}
\makeatother

\usepackage{graphicx}

\usepackage{xpinyin}
\usepackage{CJKutf8}

\usepackage[]{ACL2023}

\usepackage{times}
\usepackage{latexsym}
\usepackage[russian,english]{babel}
\newcommand{\ru}[1]{{\selectlanguage{russian}#1}\selectlanguage{english}}

\usepackage[T1]{fontenc}

\usepackage[utf8]{inputenc}

\usepackage{microtype}

\usepackage{inconsolata}
\usepackage{xspace}
\usepackage{caption}
\usepackage{subcaption}
\usepackage{enumitem}
\usepackage{quoting}
\quotingsetup{vskip=0pt}
\usepackage{url}
\usepackage{booktabs}
\usepackage{multirow}
\usepackage{graphics}
\usepackage{makecell}
\usepackage[export]{adjustbox}
\usepackage{graphbox}

\usepackage{multirow}
\usepackage{colortbl}
\usepackage{amssymb}

\usepackage[normalem]{ulem}
\useunder{\uline}{\ul}{}

\newcommand{\ignore}[1]{}

\newcommand{\parheader}[1]{{\bf \smallskip \noindent #1.}}
\newcommand{\parheaderNoDot}[1]{{\bf \smallskip \noindent #1\xspace}}

\newcommand{\greencolor}{\textcolor[HTML]{55a868}{green}\xspace}
\newcommand{\redcolor}{\textcolor[HTML]{c44e52}{red}\xspace}
\newcommand{\yellowcolor}{\textcolor[HTML]{dd8452}{yellow}\xspace}
\newcommand{\bluecolor}{\textcolor[HTML]{4c72b0}{blue}\xspace}
\newcommand{\browncolor}{\textcolor[HTML]{98623c}{brown}\xspace}

\newcommand{\blue}[1]{\textcolor[HTML]{4c72b0}{#1}\xspace}
\newcommand{\orange}[1]{\textcolor[HTML]{e18727}{#1}\xspace}

\newcommand{\cross}[0]{\textcolor[HTML]{DA3F32}{\bf $\times$}\xspace}
\newcommand{\tick}[0]{\textcolor[HTML]{398D64}{\bf \checkmark}\xspace}

\newcommand{\kmeans}[0]{K-Means\xspace}

\newcommand{\nLanguage}[0]{{28}\xspace}
\newcommand{\tydiqagoldp}[0]{TyDiQA-GoldP\xspace}
\newcommand{\tydiqa}[0]{TyDiQA\xspace}
\newcommand{\masakhaner}[0]{MasakhaNER\xspace}
\newcommand{\xnli}[0]{XNLI\xspace}
\newcommand{\miracl}[0]{MIRACL\xspace}
\newcommand{\firstk}[0]{First-$k$\xspace}

\title{
\raisebox{-1.4em}[0pt][0pt]{\includegraphics[height=2.7em]{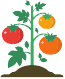}}
Tomato, Tomahto, Tomate:\
Do Multilingual Language Models \\
~~~~~~~~~Understand Based on Subword-Level Semantic Concepts?
} 

\author{
Crystina Zhang$^{1}$\thanks{~~Work is done while Crystina Zhang was a student researcher at Google DeepMind.},
Jing Lu$^{2}$, Vinh Q. Tran$^{2}$, Tal Schuster$^{2}$, Donald Metzler$^{2}$, Jimmy Lin$^{1}$ \\[1ex]
$^{1}$University of Waterloo~~~$^{2}$Google DeepMind \\[1ex]
}

\begin{document}
\maketitle

\begin{abstract}
Human understanding of text depends on general semantic concepts of words rather than their superficial forms. To what extent does our human intuition transfer to language models? In this work, we study the degree to which current multilingual language models (mLMs) understand based on subword-level semantic concepts. To this end, we form ``semantic tokens'' by merging the semantically similar subwords and their embeddings, and evaluate the updated mLMs on five heterogeneous multilingual downstream tasks. Results show that the general shared semantics could get the \mbox{models} a long way in making the predictions on mLMs with different tokenizers and model~sizes. Inspections of the grouped subwords show that they exhibit a wide range of semantic similarities, including synonyms and translations across many languages and scripts.  Lastly, we find that the zero-shot results with semantic tokens are on par with or even better than the original models on certain classification tasks, suggesting that the shared subword-level semantics may serve as the anchors for cross-lingual transfer.
\end{abstract}

\section{Introduction}
\begin{figure}
    \centering

    \includegraphics[width=\columnwidth,clip,trim=0 0mm 0 0]{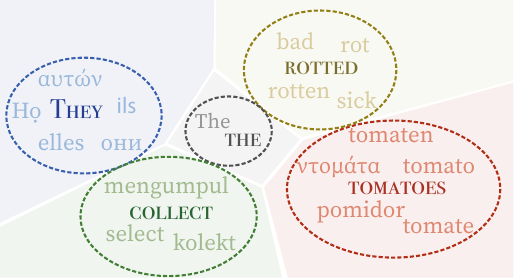}

    \caption{
        Words with similar meanings or inflectional changes fall under semantic concepts, as indicated by the colors.
        The sentence ``They collected the rotted tomatoes.'' is from XNLI~\cite{conneau-etal-2018-xnli}.
    }
    \vspace{-0.3em}
    \label{fig:teaser}
\end{figure}

Human understanding of text depends on general semantic concepts of words that are robust to their superficial forms (\autoref{fig:teaser}).
The most obvious examples are semantically equivalent words in different languages:\ 
code-switching ``they'' to \mbox{``\ru{они}''}, or ``tomatoes'' to ``tomate''.\footnote{~``\ru{они}'': ``they'' in Russian; ``tomate'': ``tomato'' in Spanish}
The sentence meaning is preserved for people who understand both languages.
The robustness in understanding may also apply to some of the inflectional changes:
swapping ``tomatoes'' for ``tomato'',
while losing the plural form, still conveys the overall sentence meaning.
Finally,
even the words are replaced with ones that are only vaguely related in semantics, e.g., from ``rotted'' to ``bad'',
resulting in a phrase that is less precise,
the sentence meaning could still be interpreted and in line with the original sentence.
While the definition and representation of concept have not been universally agreed upon~\cite{jackendoff1988conceptual,gabora2008toward,goddard2013words,gardenfors2014geometry,fumagalli2019representation,sajjad-etal-2022-analyzing},
and their alignment varies across languages depending on domain and cultural context~\cite{thompson2020cultural}, 
it is generally accepted that concepts underlying words could be shared or even universal across languages~\cite{bundy1984semantic,goddard2013words} and have a pivotal role in many aspects of language understanding~\cite{murphy2004big,fumagalli2019representation}.

\begin{figure*}
    \centering
    \includegraphics[width=\textwidth]{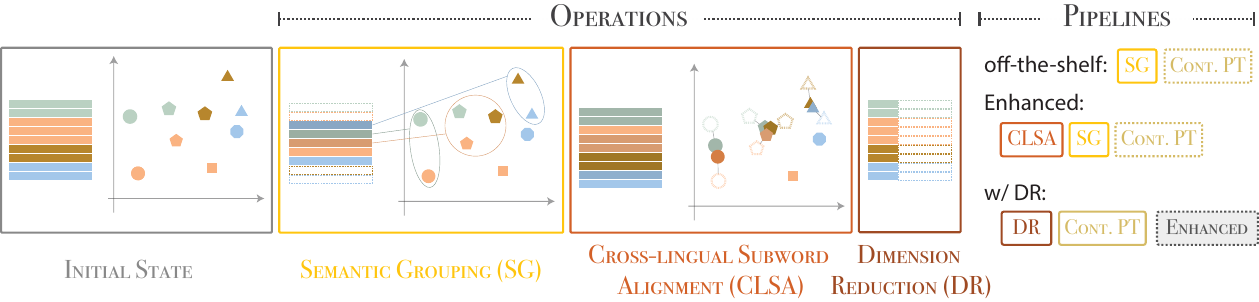}

    \caption{
        Illustrations of the operations that modify the word embeddings and the pipelines composed of these~operations.
        ``Cont. Pt'': Continual Pretraining, whose illustration is skipped due to its prevalence. 
        The colored rows represent the embeddings of the subwords and the coordinates depict their spatial distances.
        The shapes indicate the underlying semantics and the colors indicate the languages of the subwords.
        Under the ``\textsc{Pipelines}'',
        the solid boxes denote the required operations that are core to the pipeline or essential to the model functionality;
        the dashed boxes denote the optional operations that are only to provide additional improvement. 
        Better viewed in color.
    }
    \label{fig:operations}
\end{figure*}

On the other hand,
language models learn distinct embedding vectors for these subwords,
which share similar underlying semantics yet may not share similar context.
We thus ask:\
{\it To what degree do current multilingual language models (mLMs) understand based on subword-level semantic concepts?}
Based on the existing mLMs,
we form ``semantic tokens'' by grouping the subwords based on the similarity of their word embeddings,
which henceforth share the same ``semantic embedding''.\footnote{~We understand that the formed semantic tokens are not perfect and may contain subwords of loosely related or unrelated meanings.
On one hand, we provide inspection of the formed semantic tokens in \autoref{fig:inspectionCluster} and \ref{fig:ap:inspection}, 
which suggest that the grouped tokens could reflect coherent semantics.
On the other hand, we consider our results as a lower bound, 
where techniques that form more accurate semantic tokens are likely to further improve the downstream effectiveness.
}
The updated mLMs are evaluated on five downstream tasks, which cover thirty languages in total and include classification and embedding tasks in different granularities.

We find that with a small number of semantic tokens and their embeddings,
the mLMs can preserve most of the downstream effectiveness:\
semantic tokens in 5\% of the original vocabulary size achieve 90\% of the effectiveness on classification tasks, 
and 20\% of the semantic tokens achieve over 85\% effectiveness on the embedding tasks.
These findings suggest that while nuances exist in the meaning of each subword,
the general semantics representations could get the models a long way in prediction-making.

Next, we eliminate the confounding factor of embedding size:\ 
While forming semantic tokens, the number of the word embedding parameters is also reduced.
Does the change in effectiveness stem from changes in subwords or reduced embedding parameters?
We thus apply the semantic grouping to the word embeddings with reduced parameter size via truncating the embedding dimension,
finding that the same observation persists even when the parameters could not be further reduced via dimension reduction.
This suggests that the above results are not confounded by the embedding size but rather are a result of the semantic grouping.

Additional experiments suggest that the findings generalize to mLMs with different tokenizers and model sizes.
Inspection shows that the grouped subwords indeed exhibit a wide range of semantic similarities:\ 
numbers, punctuation, synonyms, and translations across multiple languages under different scripts.
Lastly,
we find that the zero-shot results on certain classification tasks
with semantic tokens are on par with or even better than the original models,
suggesting that the shared subword-level semantics may serve as the transfer anchors for cross-lingual generalization.

Our contributions are as follows:
{(1)} We find that mLMs can preserve a majority of the downstream effectiveness with a small number of shared subword-level semantics 
 (Section~\ref{sec:results:main}, \ref{sec:results:dimensionReduction}).
{(2)} We show that the findings are general across mLMs with different tokenizers, model sizes, and other aspects 
(Section~\ref{sec:results:backbones}).
{(3)}~Inspection reveals that the grouped subwords exhibit a wide range of semantic similarities 
(Section~\ref{sec:analysis}).
{(4)} The zero-shot results suggest that the shared subword-level semantics may serve as transfer anchors for cross-lingual generalization (Section~\ref{sec:results:zero-shot}).

\begingroup
\setlength\cmidrulewidth{0.01pt}
\begin{table*}[t]
\centering
\resizebox{\textwidth}{!}{
    \begin{tabular}{llllrp{0.47\linewidth}}
    \toprule
    \textbf{Dataset Name} & \textbf{Task Name} & \textbf{Task Type} & \textbf{Granularity} & \multicolumn{1}{l}{\textbf{\# L.}} & \textbf{Languages} \\
    \midrule
    MasakhaNER & NER & classification & word-level & 10 & am, ha, ig, rw, lg, sw, wo, yo, luo, pcm \\
    \cmidrule{1-6}
    XNLI & NLI & classification & sentence-level & 15 & ar, bg, de, el, en, es, fr, hi, ru, sw, th, tr, ur, vi, zh \\
    \cmidrule{1-6}
    TyDi QA & QA & classification & sentence-level & 11 & ar bn, en, fi, d, ja, ko, ru, sw, te, th \\
    \cmidrule{1-6}
    \multirow{2}{*}{MIRACL} & P Retrieval & classification & \multirow{2}{*}{passage-level} & \multirow{2}{*}{18} & \multirow{2}{1\linewidth}{ar, bn, de, en, es, fa, fi, fr, hi, id, ja, ko, ru, sw, te, th, yo, zh} \\
     & P Reranking & embedding &  &  &  \\
    \bottomrule
    \end{tabular}
}
\caption{
Downstream Tasks and Datasets.
Granularity suggests the information level required from input data to perform the task.
{
``P Retrieval'': Passage Retrieval;
``P Reranking'': Passage Reranking.}
}
\vspace{-0.5em}
\label{tab:datasets}
\end{table*}
\endgroup
\section{Operations and Pipelines}
The experimental setting in this paper can be categorized into several pipelines composed of multiple independent operations,
centering on semantic grouping.
All operations, except for the continual pretraining, only affect the vocabulary $V$ and word embeddings $E \in R^{(|V|, D)}$, where $|V|$ is the vocabulary size and $D$ is the initial word embeddings dimension.
\autoref{fig:operations} illustrates all operations except the continual pretraining due to its prevalence,
as well as the pipelines composed of the operations.

\subsection{Semantic Grouping (SG)}
\label{sec:method:sg}

Given the vocabulary $V$ and its word embeddings $E$, 
multiple semantically similar subwords are grouped into a single ``semantic token'' and henceforth share the same ``semantic embedding''. 
The semantic embedding is then initialized by the averaged embeddings of the grouped words.
That is, after the grouping, the updated LM has a new vocabulary $V'$ composed of semantic tokens and new word embeddings $E^{'}_{vocab} \in R^{(|V'|, D)}$, where $|V'| < |V|$.
We define the grouping ratio as $r_{G} = {|V'|}/{|V|}$.

\parheader{\kmeans}
Subwords are grouped via \kmeans based on the cosine distance of their word embeddings.
We choose \kmeans due to its flexibility in the produced number of groups $|V'|$,
and use the cosine distance as experiments show that it has better performance, especially at high word embeddings dimensions (Ap.~\ref{ap:ablation:distance}).
The inspections of the groups show that they could reflect coherent semantics (\autoref{fig:inspectionCluster} and \ref{fig:ap:inspection}).
In this work, we set $|V'|$ to correspond to a grouping ratio $r_{G} \in \{5\%, 10\%, 20\%, 40\%\}$.\footnote{~We also investigate grouping based on bilingual lexical mappings in pilot studies.
See results comparison in Ap.~\ref{ap:sec:bilingual}.
}

\parheader{First-$k$}
As the baseline, we keep only the \mbox{first-$k$} emerged subwords while training the tokenizer, where $k=|V'|$.
In this way, the size of the vocabulary and word embedding vectors remains the same as in the corresponding semantically grouped models,
yet each embedding corresponds to only a single subword as the original LM.

\subsection{Cross-lingual Subword Alignment (CLSA)}
\label{sec:method:clsa}
Other than clustering on the off-the-shelf word embedding,
we investigate manually aligning the embeddings of subwords across different languages.
Specifically, we gather cross-lingual word pairs from bilingual dictionaries
(i.e., MUSE;~\citealp{conneau2017word}; PanLex\footnote{~\url{https://panlex.org/}})
and concept lists of multilingual words (i.e., Concepticon;~\citealp{list-etal-2016-concepticon}; ColexNet;~\citealp{liu-etal-2023-crosslingual-transfer}).
We only preserve the words that are tokenized into a single subword.
Then, the word embeddings are trained using InfoNCE loss~\cite{oord2018representation} with in-batch negatives.
Note that only the parameters of word embeddings are used and updated in this operation,
while the rest of the model remains untouched.
The training configurations of the CLSA operation are provided in Ap.~\ref{fig:ap:alignment-source},
as well as the ablations on the source datasets used for CLSA training.

\subsection{Dimension Reduction (DR)}
\label{sec:methods:dimensionReduction}
Section~\ref{sec:results:dimensionReduction} involves experiments that reduce the dimension of word embeddings,
where 
we simply {\it remove} the final $D - d$ dimensions of the word embeddings to form the new word embeddings $E^{'}_{d} \in R^{(|V|, d)}$ ($d < D$),
and pad each embedding vector with zeros on the fly.
Note that the positional and token-type embeddings are not affected by this operation.
While there are alternative options to reduce the word embedding parameters,
we adopt DR for its simplicity and to minimize changes in model architecture.
Due to its great modifications to the word embeddings,
this operation is always followed by continual pretraining (Section~\ref{sec:method:contPt}).

\subsection{Continual Pretraining}
\label{sec:method:contPt}

The above operations may lead to a potential mismatch between the word embeddings and the rest of the model parameters.
To address the mismatch,
we continually pretrain the entire LM using Masked Language Modeling (MLM) objectives to align the updated embeddings and language model parameters~\cite{devlin-etal-2019-bert}.
All continual pretraining uses Wikipedia data of \nLanguage languages,\footnote{~ar, bg, bn, de, el, en, es, fa, fi, fr, ha, hi, id, ig, ja, ko, lg, ru, rw, sw, te, th, tr, ur, vi, wo, yo, zh
}
where the languages are selected based on the coverage of downstream tasks.
Details on the configurations are provided in Ap.~\ref{ap:config}.

\begin{figure*}[t]
    \centering
        \begin{subfigure}[t]{0.195\textwidth}
            \includegraphics[width=1\columnwidth,trim={0.3cm 0cm 0cm 0},clip]{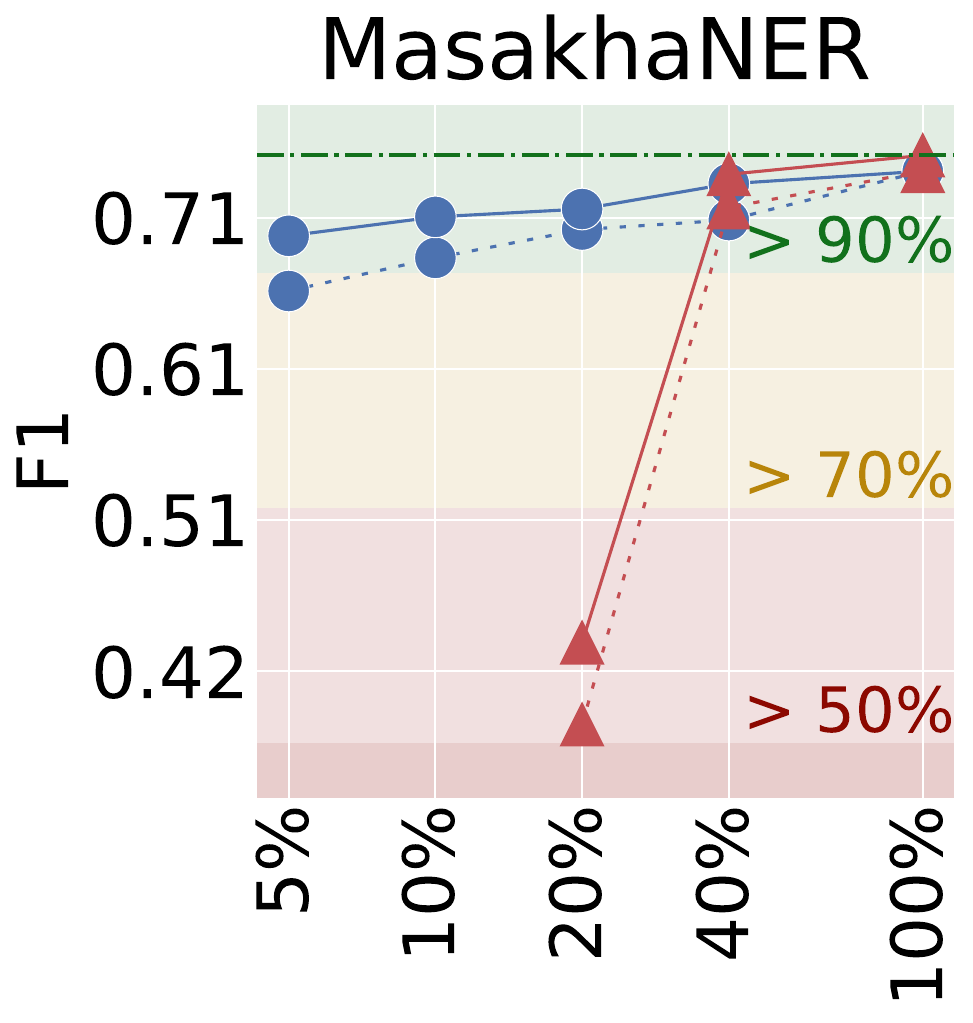}
        \end{subfigure}
        \hfill
        \begin{subfigure}[t]{0.195\textwidth}
            \includegraphics[width=1\columnwidth,trim={0.3cm 0cm 0cm 0},clip]{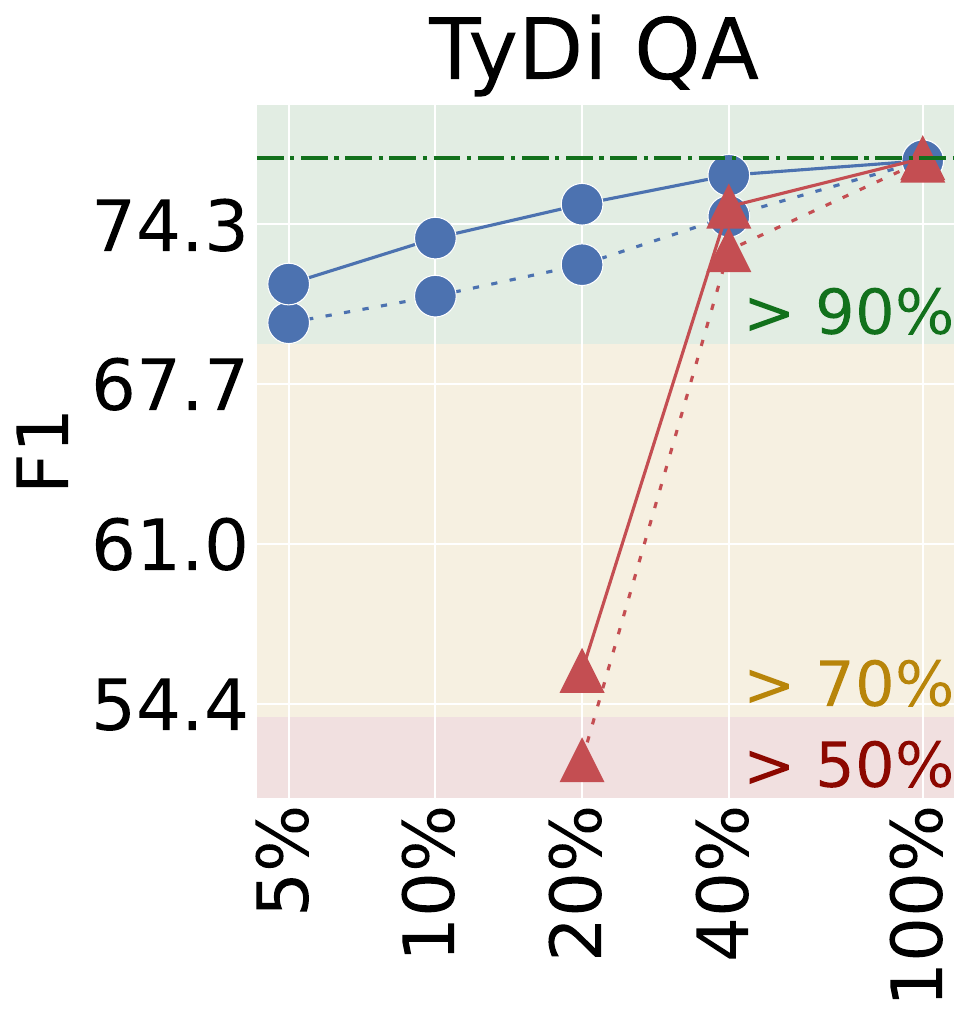}
        \end{subfigure}
        \hfill
        \begin{subfigure}[t]{0.195\textwidth}
            \includegraphics[width=1\columnwidth,trim={0.3cm 0cm 0cm 0},clip]{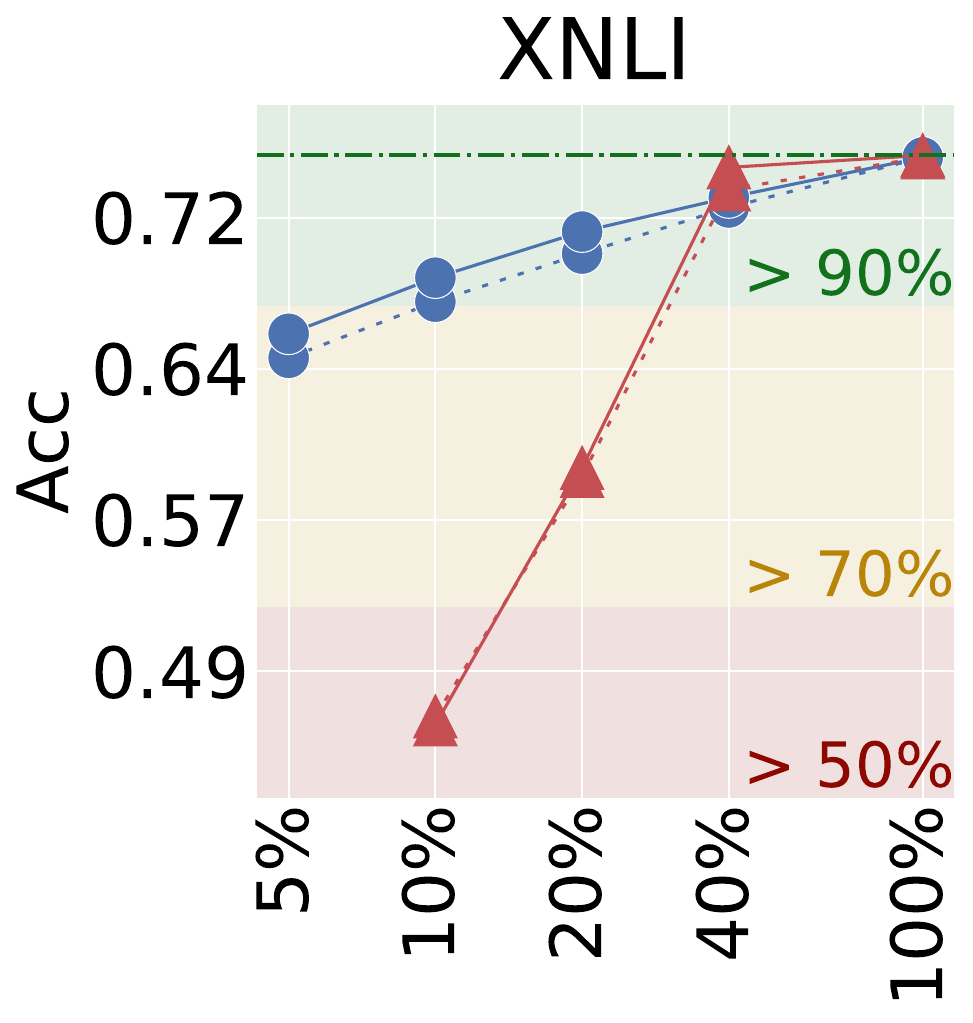}
        \end{subfigure}
        \hfill
        \begin{subfigure}[t]{0.195\textwidth}
            \includegraphics[width=1\columnwidth,trim={0.3cm 0cm 0cm 0},clip,right]{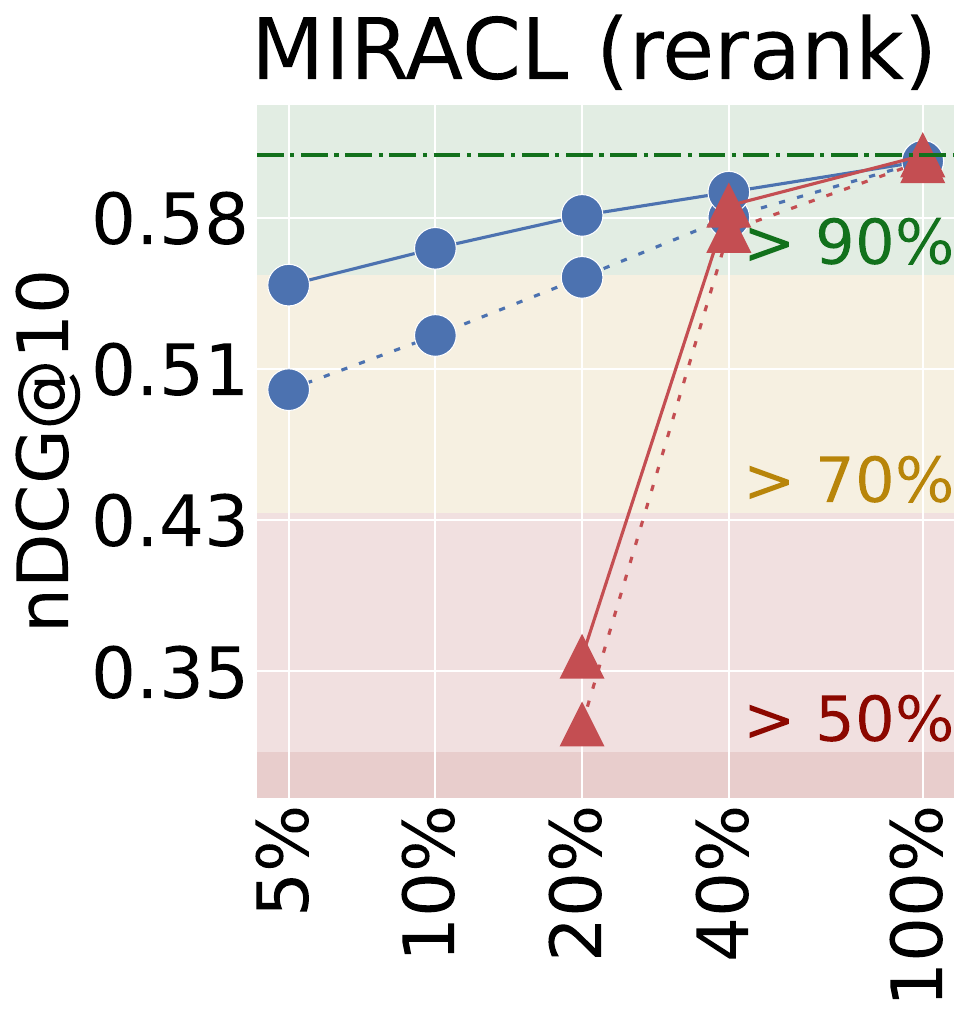}
        \end{subfigure}
        \hfill 
        \begin{subfigure}[t]{0.195\textwidth}
            \includegraphics[width=1\columnwidth,trim={0.3cm 0cm 0cm 0},clip,left]{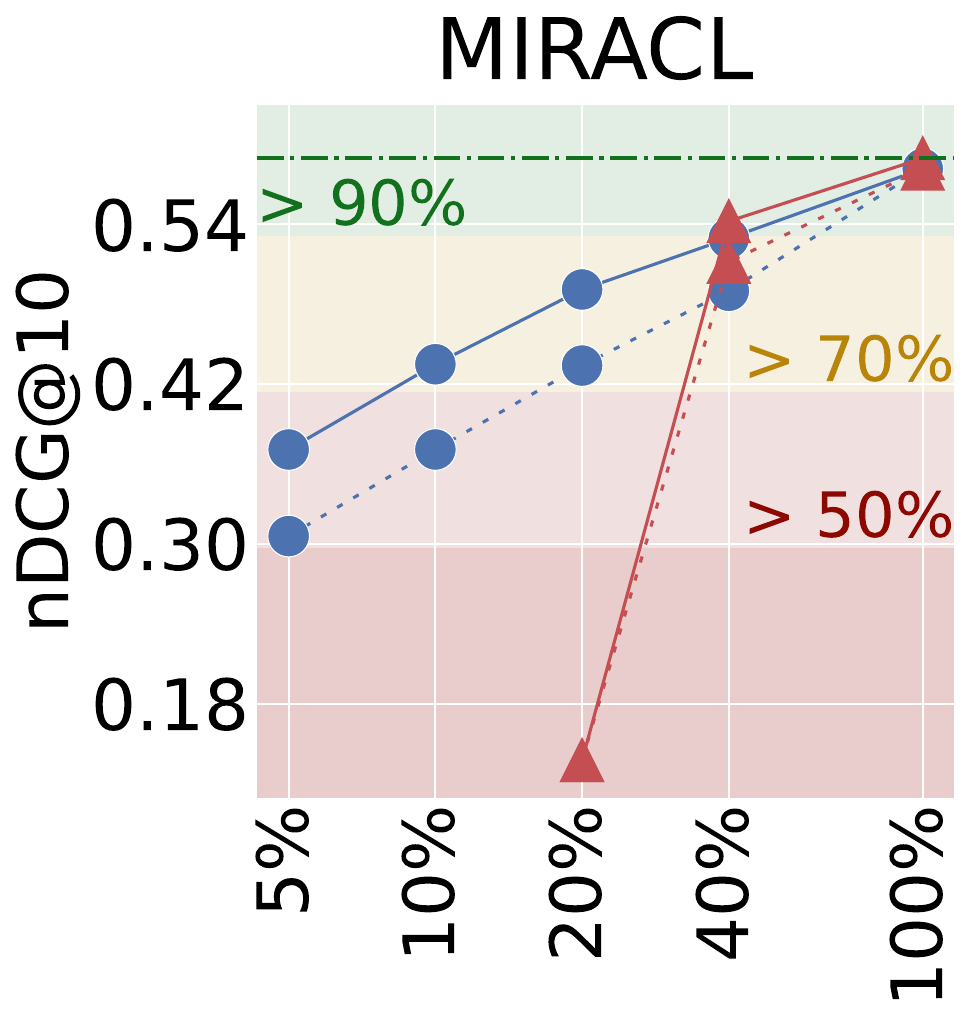}
        \end{subfigure}

        \begin{subfigure}[t]{0.75\textwidth}
            \includegraphics[width=0.95\columnwidth,trim={0cm 0cm 0cm 0},clip]{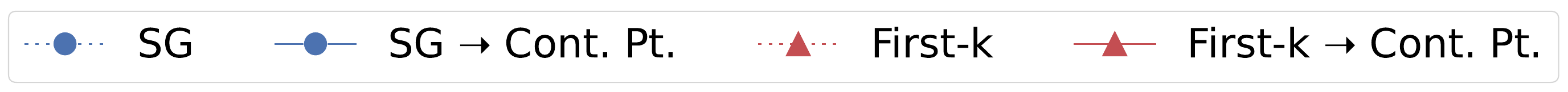}
        \end{subfigure}

    \caption{
        Results of mBERT with 
        vocabulary and embeddings after semantic grouping (SG) or simply reduced size \mbox{(\firstk)}.
        {\bf x-axis}: the grouping ratio $r_G$ in log scale.
        The background colors indicate the relative performance to the oracle results, i.e., continual pretrained mBERT with full vocabulary, indicated by the green dashed lines on top.
        \greencolor: >90\%, \yellowcolor: 70\%--90\%, \redcolor: 50\%--70\%. 
        The scores of First-$k$ at $r_G = \{5\%, 10\%\}$ are skipped in the figures as they greatly skew the y-axis.
    }
    \label{fig:compareRandom768}
\end{figure*}

\begin{figure*}[t]
    \vspace{-0.5em}
    \centering
        \begin{subfigure}[t]{0.195\textwidth}
            \includegraphics[width=1\columnwidth,trim={0.3cm 0cm 0cm 0},clip]{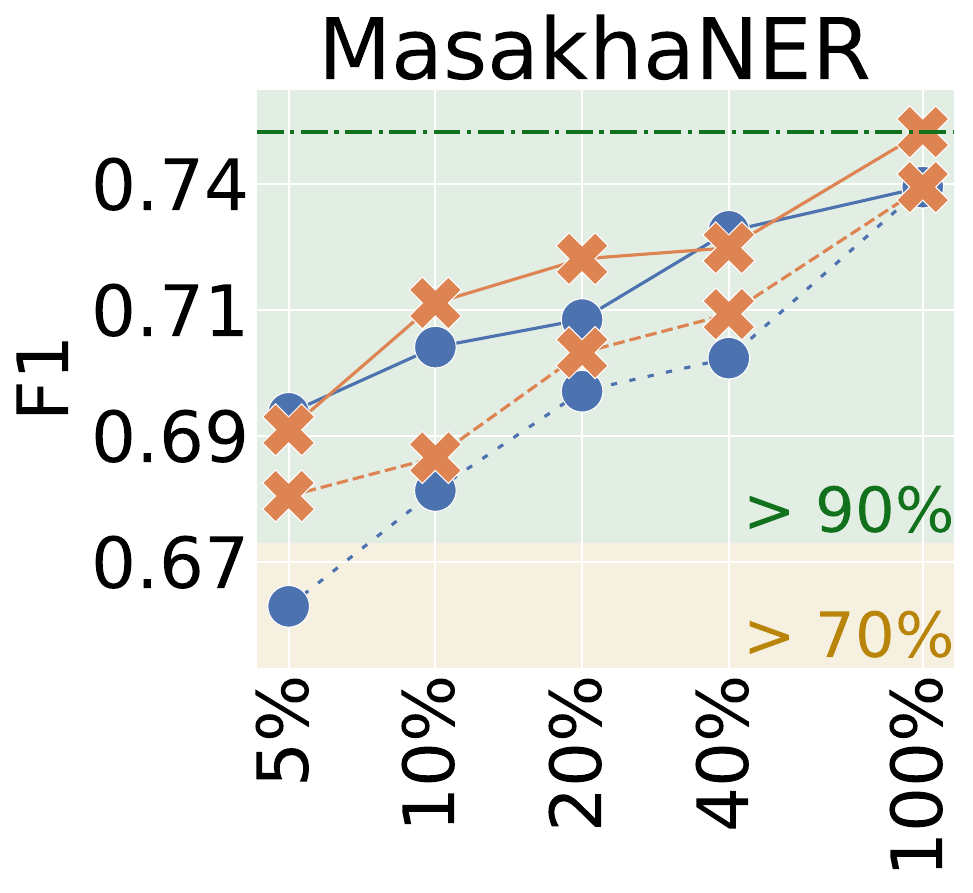}
        \end{subfigure}
        \hfill
        \begin{subfigure}[t]{0.195\textwidth}
            \includegraphics[width=1\columnwidth,trim={0.3cm 0cm 0cm 0},clip]{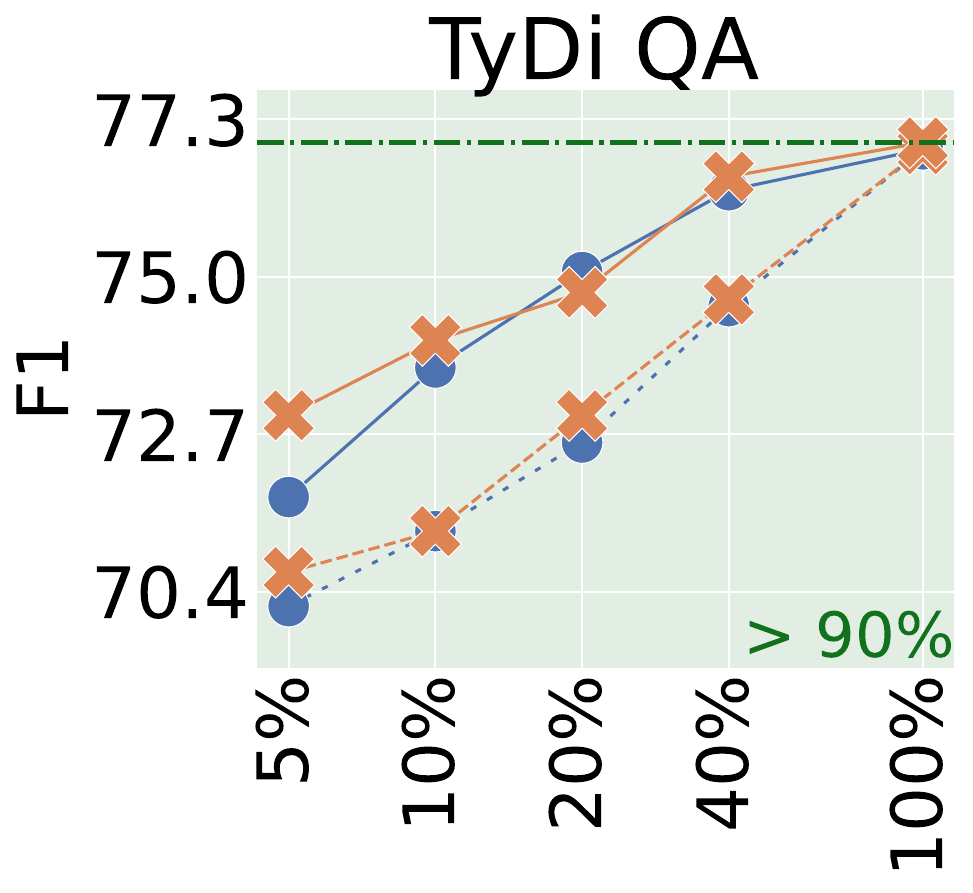}
        \end{subfigure}
        \hfill
        \begin{subfigure}[t]{0.195\textwidth}
            \includegraphics[width=1\columnwidth,trim={0.3cm 0cm 0cm 0},clip]{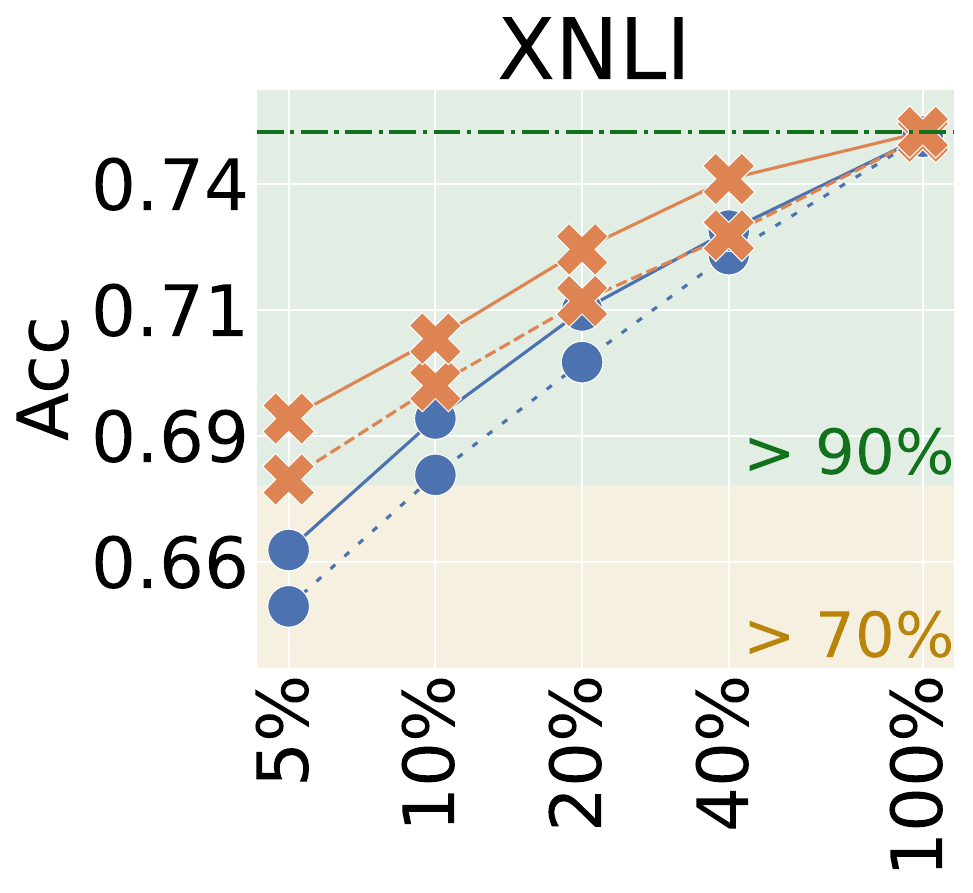}
        \end{subfigure}
        \hfill
        \begin{subfigure}[t]{0.195\textwidth}
            \includegraphics[width=1\columnwidth,trim={0.3cm 0cm 0cm 0},clip,right]{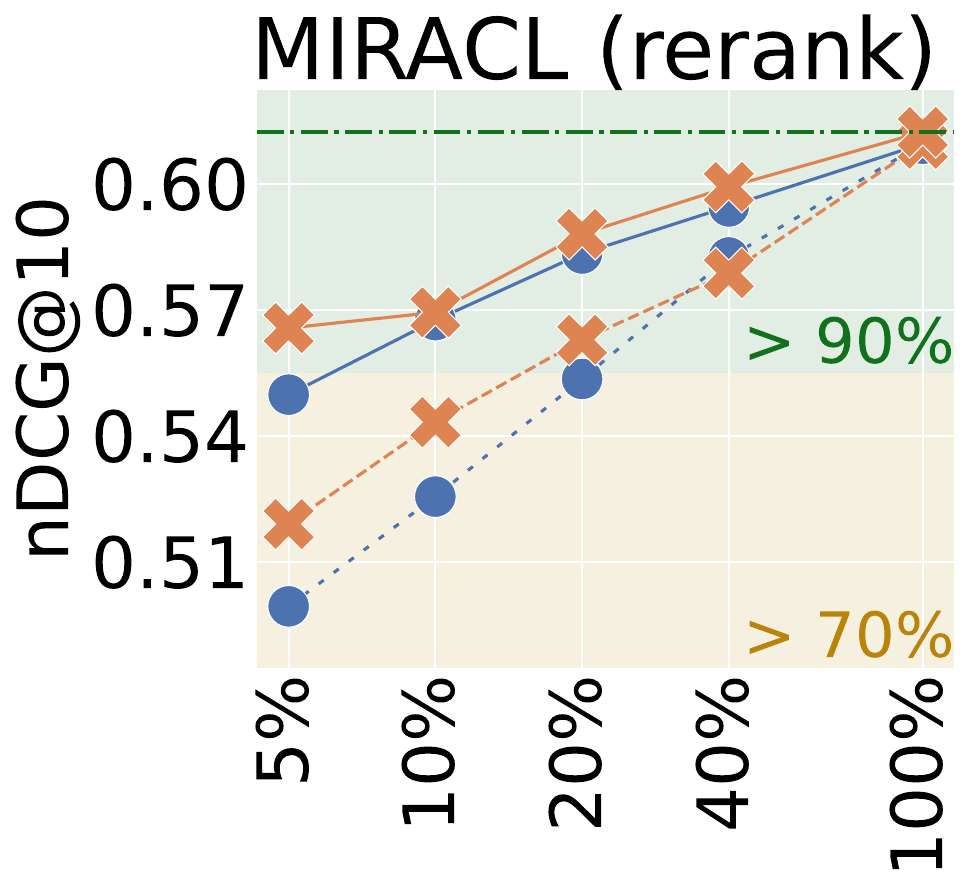}
        \end{subfigure}
        \hfill 
        \begin{subfigure}[t]{0.195\textwidth}
            \includegraphics[width=1\columnwidth,trim={0.3cm 0cm 0cm 0},clip,left]{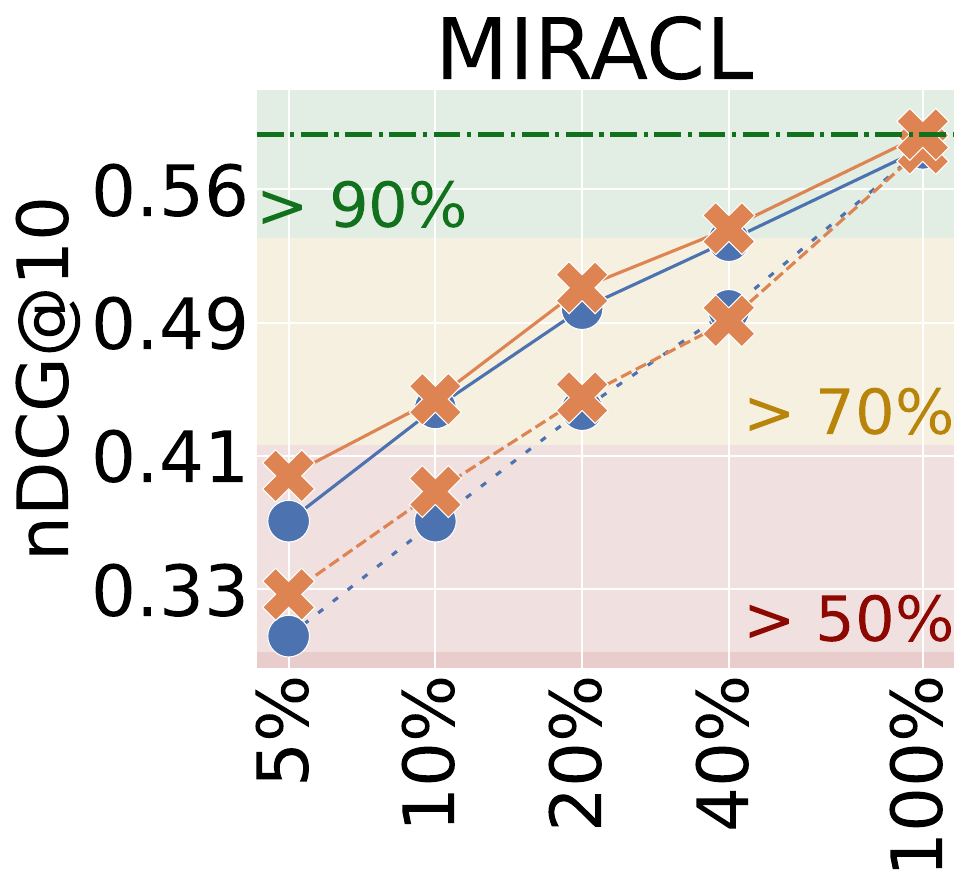}
        \end{subfigure}

        \begin{subfigure}[t]{0.8\textwidth}
            \includegraphics[width=0.95\columnwidth,trim={0cm 0cm 0cm 0},clip]{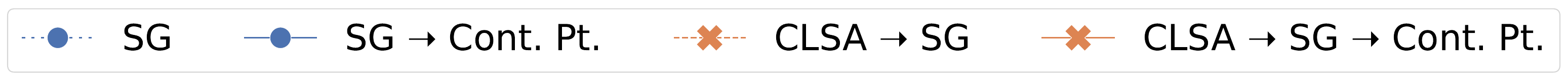}
        \end{subfigure}

    \vspace{-0.5em}
    \caption{
        Results of mBERT after semantic grouping (SG) \orange{with} and \blue{without} applying cross-lingual subword alignment (CLSA).
        Background colors design is identical to \autoref{fig:compareRandom768}.
    }
    \label{fig:768align}
\end{figure*}

\section{Downstream Tasks Evaluation}
We evaluate all configurations on five multilingual downstream tasks,
where four are classification and one is embedding.
The classification tasks include word-level understanding, sentence-level understanding, and passage reranking,
whereas the embedding task includes passage retrieval.
All evaluation datasets cover at least 10 diverse languages.
See \autoref{tab:datasets} for details.
We hope these provide a comprehensive evaluation regarding
the task nature, model structures, and languages.

\parheaderNoDot{\masakhaner}\cite{adelani-etal-2021-masakhaner} is a named entity recognition (NER) benchmark including 10 under-represented African languages.
We use its version~1.0 in this work.

\parheaderNoDot{\tydiqagoldp}\cite{clark-etal-2020-tydi} 
is the gold passage task of \tydiqa~\cite{clark-etal-2020-tydi}, a question answering (QA) dataset that includes 11 topologically different languages.
It requires predicting correct answer based on gold answer passages.
We refer to it as \tydiqa for simplicity. 

\parheaderNoDot{\xnli}\cite{conneau-etal-2018-xnli} is a natural language inference (NLI) dataset that extends the development and test sets of MultiNLI~\cite{williams-etal-2018-broad} to 15 diverse languages.

\parheaderNoDot{\miracl}\cite{zhang-etal-2023-miracl}
is a monolingual information retrieval dataset that provides training data for 16 diverse languages and evaluation data for an additional 2 languages.
Two tasks are performed on \miracl:\ 
passage retrieval and passage reranking,
which fall under embedding and classification tasks respectively.
In the rest of this paper,
\miracl refers to the passage retrieval task and \miracl (rerank) refers to the passage reranking task.
We use the classic DPR~\cite{karpukhin-etal-2020-dense} and monoBERT~\cite{nogueira2019passagerw,nogueira2019multistagedr} models for each task.

\section{Results and Analysis}
\label{sec:results}

\subsection{Semantic Grouping (SG)}
\label{sec:results:main}

\autoref{fig:compareRandom768} illustrates the results of applying SG on mBERT~\cite{devlin-etal-2019-bert} 
across a spectrum of grouping ratios $r_G$ (x-axis).
The background colors indicate the range of relative effectiveness compared to the oracle results, which are the scores of the original mBERT with continual pretraining applied.
Each point in the figures represents the \mbox{{average}} score across all languages for the given configurations.\footnote{
~Due to space constraint,
Section~\ref{sec:results} only provides visualization of all results,
where numerical scores can be found in Ap.~\ref{ap:numerical-results}, \autoref{tab:ap:numerical-results};
We also investigate the impact on individual language in Ap.~\ref{ap:analysis:individual-lang}. While the overall trend per language is similar,
we do not observe a consistent impact over languages across different benchmarks.
}

\parheader{Semantically similar subwords could share the same embeddings to a large degree}
In~classification tasks (the left four sub-figures),
applying SG alone (the \bluecolor dashed lines) can already preserve over 90\% effectiveness on the downstream tasks with 10\% of the original vocabulary size.
After applying continual pretraining (the \bluecolor solid lines), 
the same level of effectiveness (> 90\%) can be preserved with only 5\% of the original vocabulary size.
The embedding task (the rightmost sub-figure) is comparatively more sensitive to the semantic grouping,
yet still maintains over 85\% effectiveness with 20\% of the original vocabulary size after the continual pretraining.
The different behaviors suggest that classification tasks may require only coarse semantic representations to make predictions,
while the embedding tasks require more fine-grained lexical representations to produce reasonable sentence- or passage-level representations.

\begin{figure*}[t]
    \centering
        \begin{subfigure}[t]{0.195\textwidth}
            \includegraphics[width=1\columnwidth,trim={0.3cm 0 0cm 0},clip]{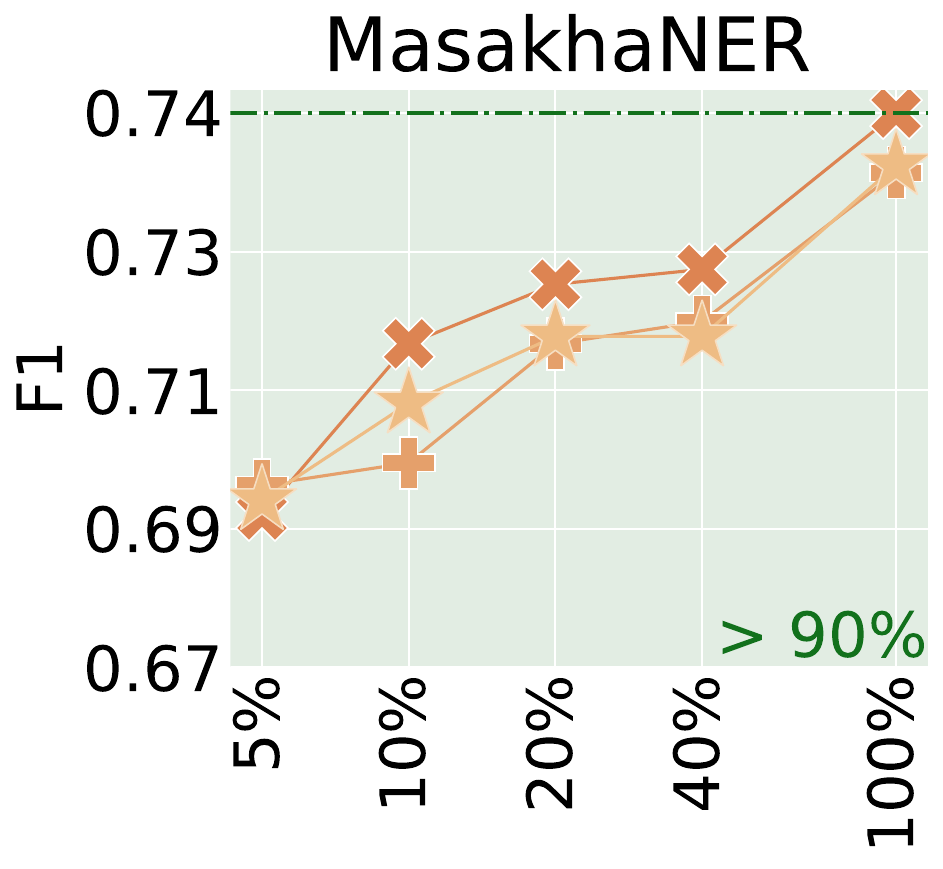}
        \end{subfigure}
        \hfill
        \begin{subfigure}[t]{0.195\textwidth}
            \includegraphics[width=1\columnwidth,trim={0.3cm 0 0cm 0},clip]{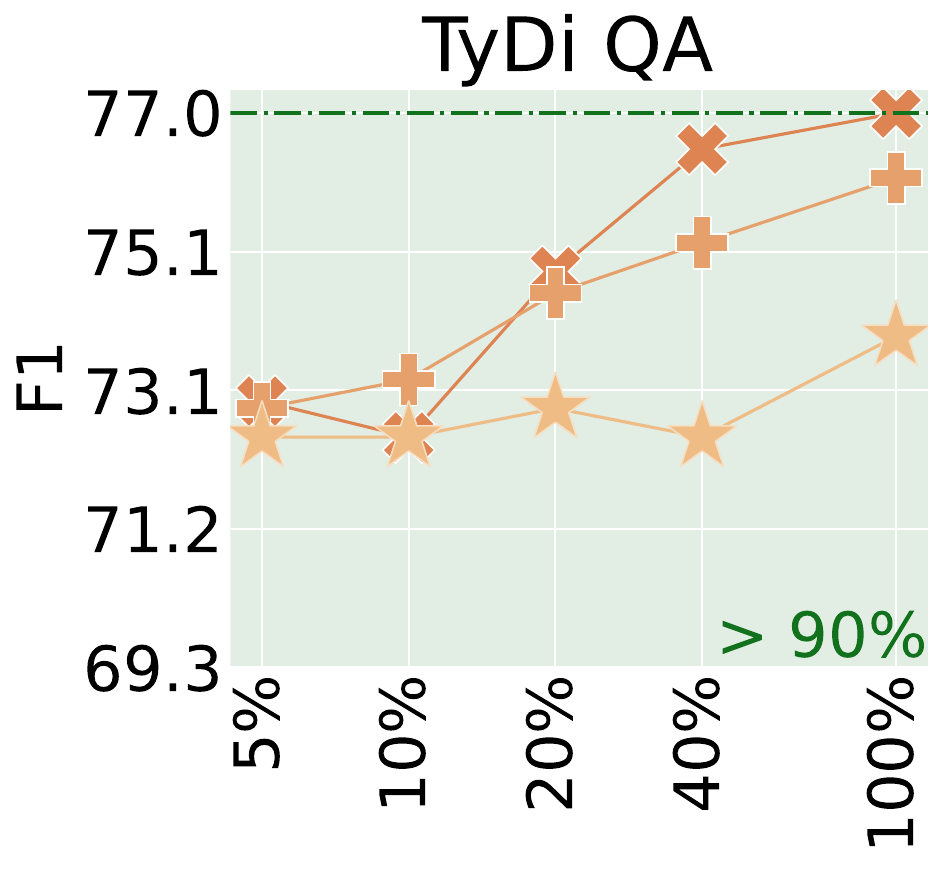}
        \end{subfigure}
        \hfill
        \begin{subfigure}[t]{0.195\textwidth}
            \includegraphics[width=1\columnwidth,trim={0.3cm 0 0cm 0},clip]{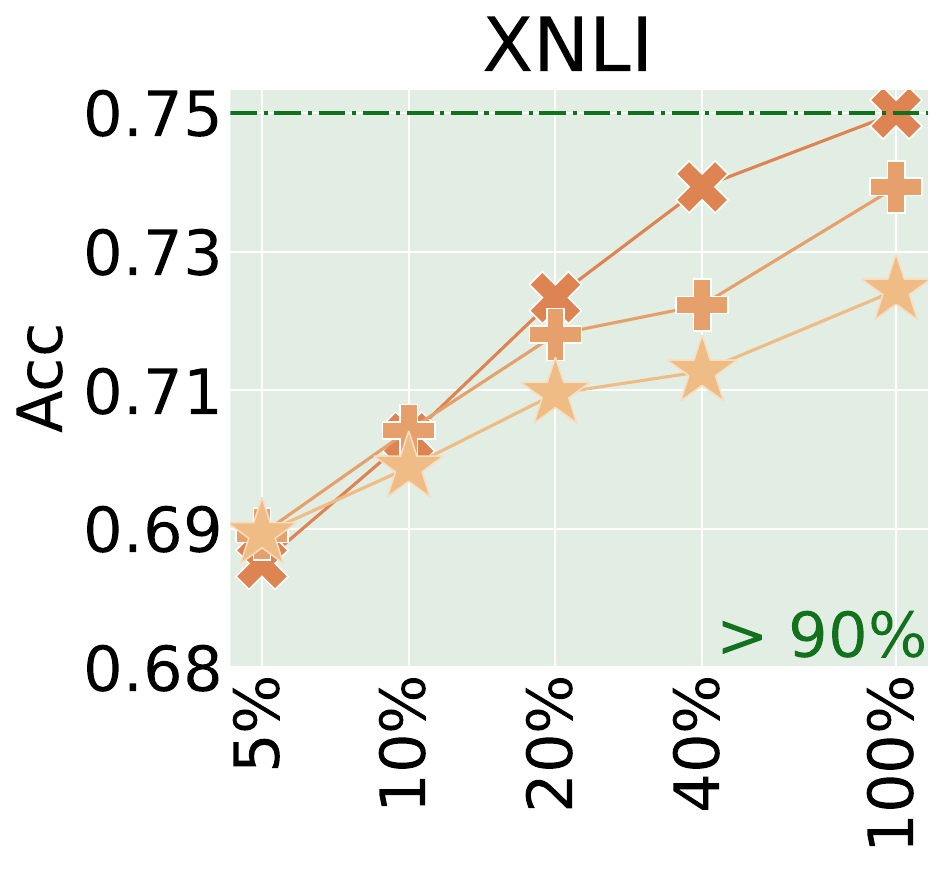}
        \end{subfigure}
        \hfill
        \begin{subfigure}[t]{0.195\textwidth}
            \includegraphics[width=1\columnwidth,trim={0.3cm 0 0cm 0},clip,right]{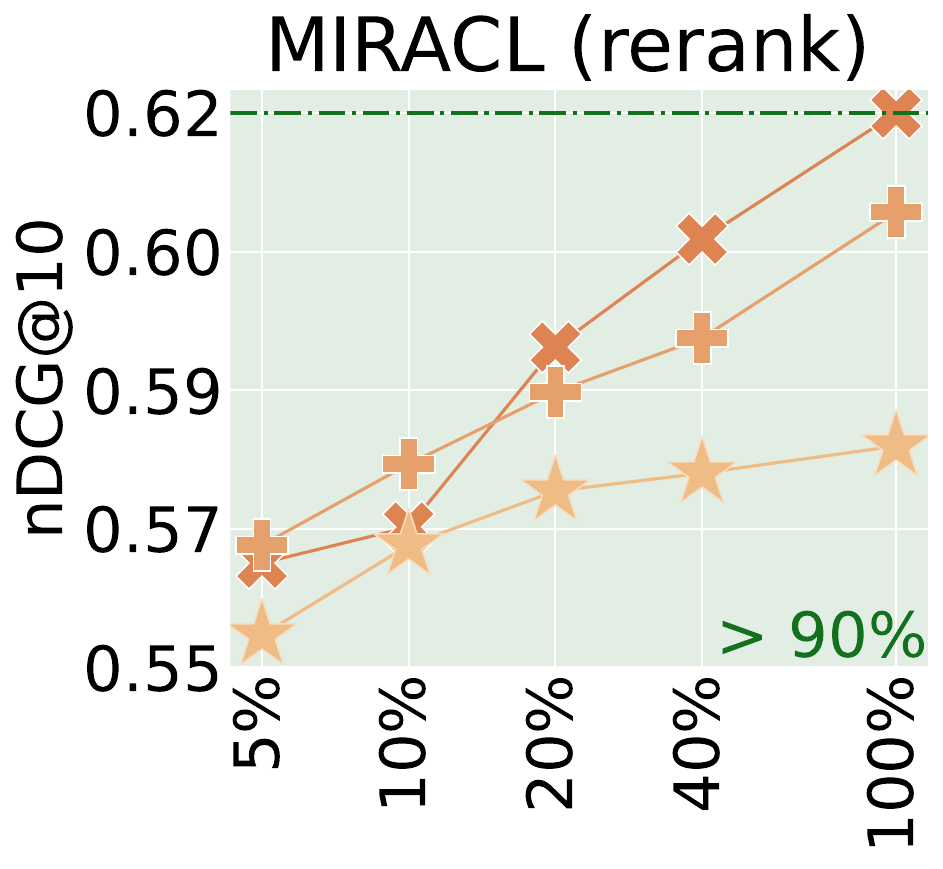}
        \end{subfigure}
        \hfill 
        \begin{subfigure}[t]{0.195\textwidth}
            \includegraphics[width=1\columnwidth,trim={0.3cm 0 0cm 0},clip,left]{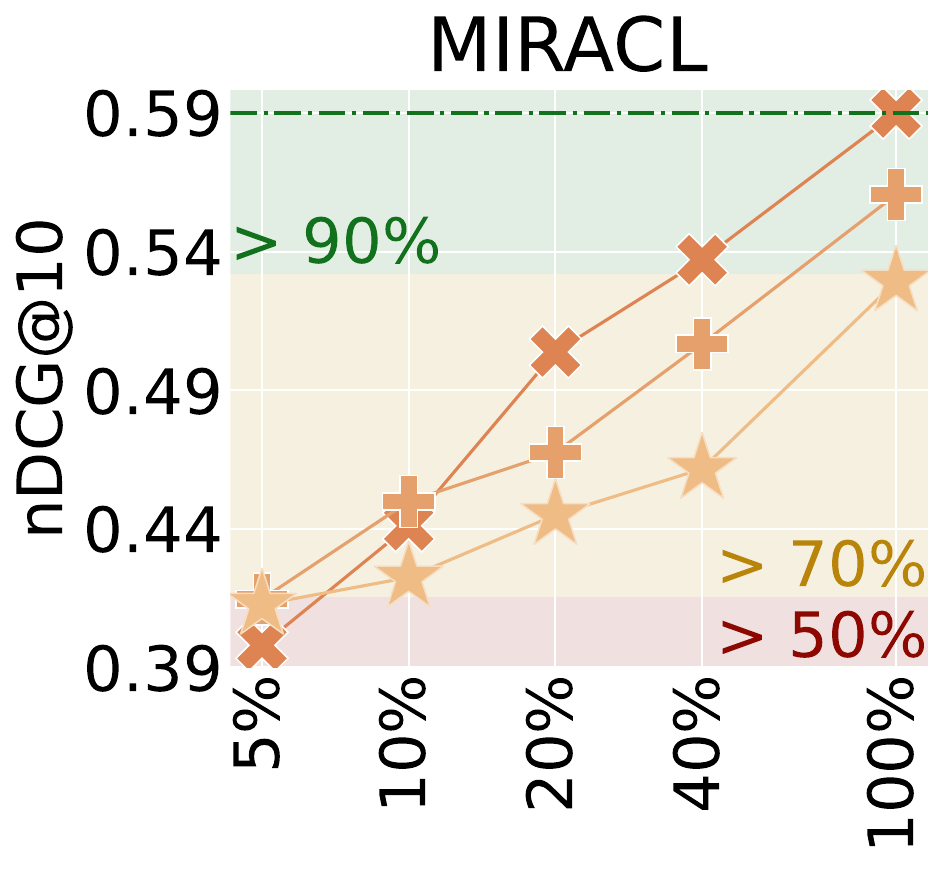}
        \end{subfigure}
        \begin{subfigure}[t]{0.43\textwidth}
            \includegraphics[width=1\columnwidth,trim={0cm 0cm 0cm 0},clip]{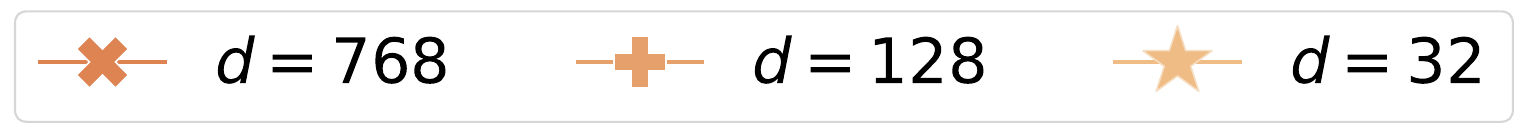}
        \end{subfigure}

    \caption{
        Results of semantic grouping with word embedding dimension $d \in \{768, 128, 32\}$, with CLSA and continual pretraining.
        Background colors design is identical to \autoref{fig:compareRandom768}.
        The relative performance of all classification tasks are higher than 90\%, thus with full \greencolor background.
    }
    \label{fig:3dim}
\end{figure*}

\begin{figure*}[t]
    \centering
    \begin{subfigure}[t]{0.195\textwidth}
        \includegraphics[width=1\columnwidth,trim={0 0cm 0cm 0},clip]{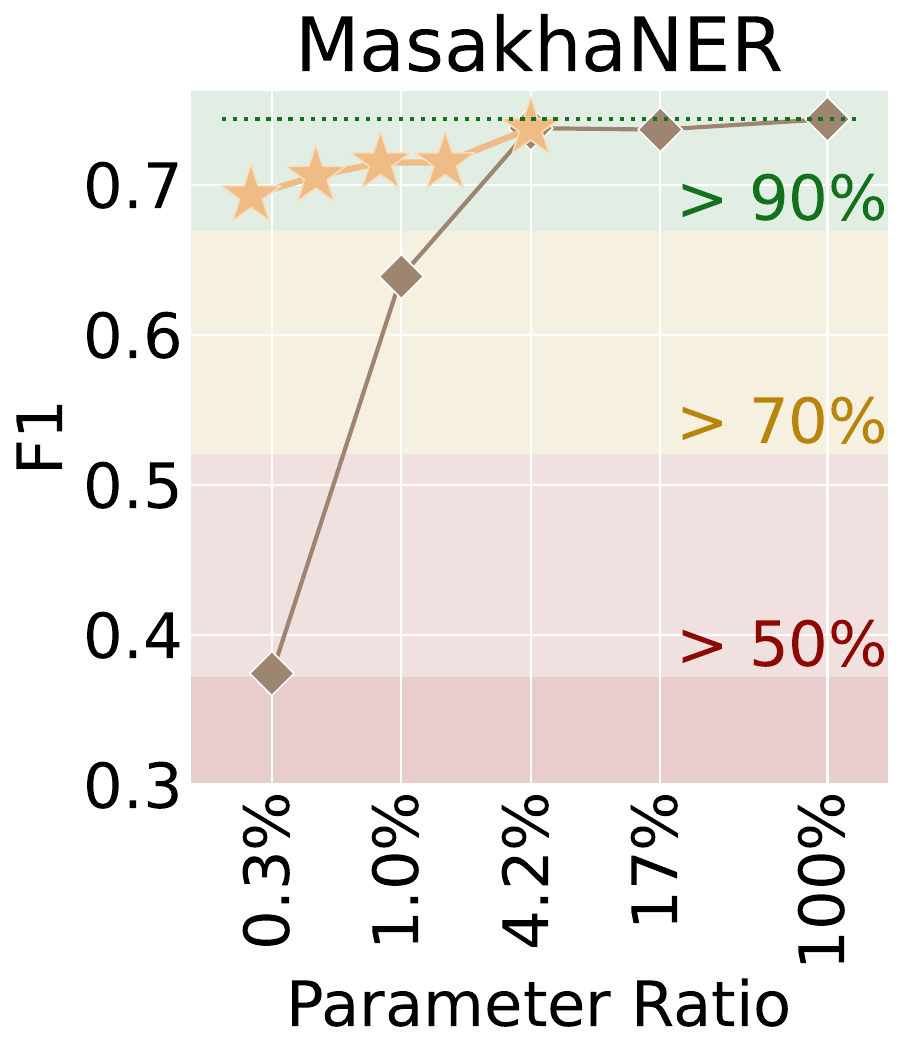}
    \end{subfigure}
    \hfill
    \begin{subfigure}[t]{0.195\textwidth}
        \includegraphics[width=1\columnwidth,trim={0 0cm 0cm 0},clip]{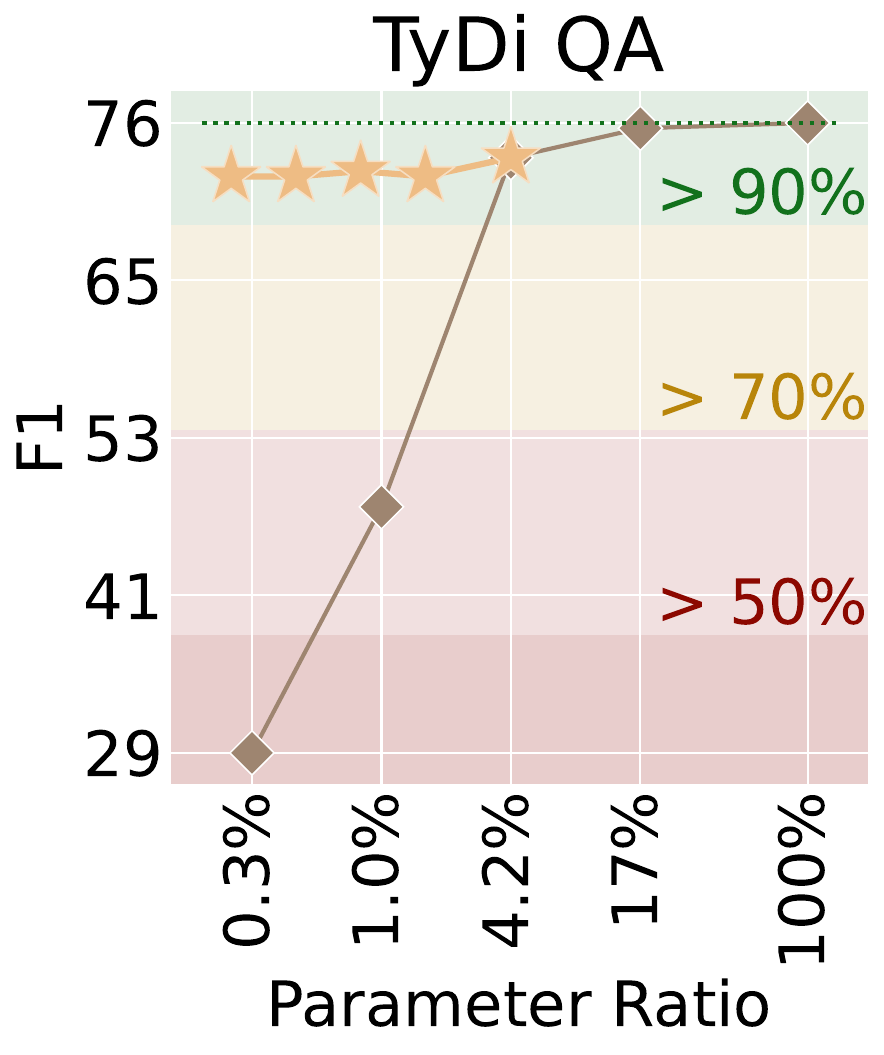}
    \end{subfigure}
    \hfill
    \begin{subfigure}[t]{0.195\textwidth}
        \includegraphics[width=1\columnwidth,trim={0 0cm 0cm 0},clip]{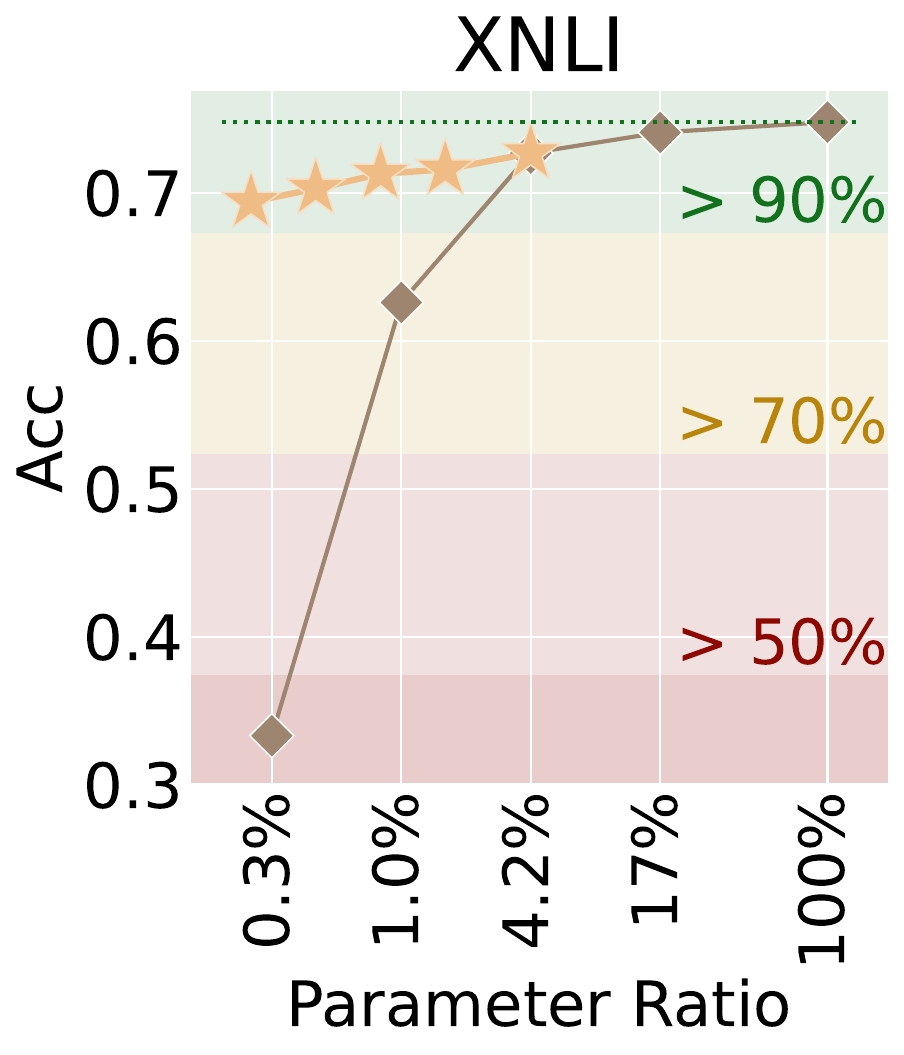}
    \end{subfigure}
    \hfill    
    \begin{subfigure}[t]{0.195\textwidth}
        \includegraphics[width=1\columnwidth,trim={0 0cm 0cm 0},clip]{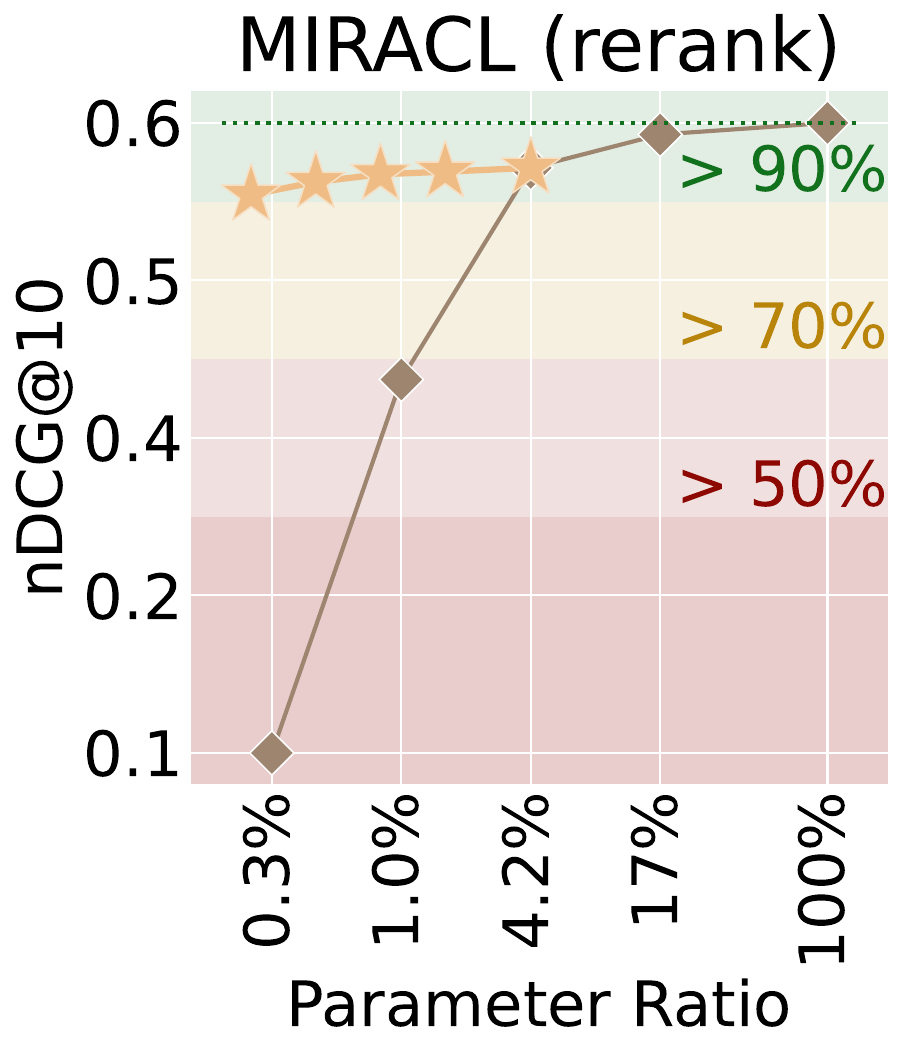}
    \end{subfigure}
    \hfill 
    \begin{subfigure}[t]{0.195\textwidth}
        \includegraphics[width=1\columnwidth,trim={0 0cm 0cm 0},clip]{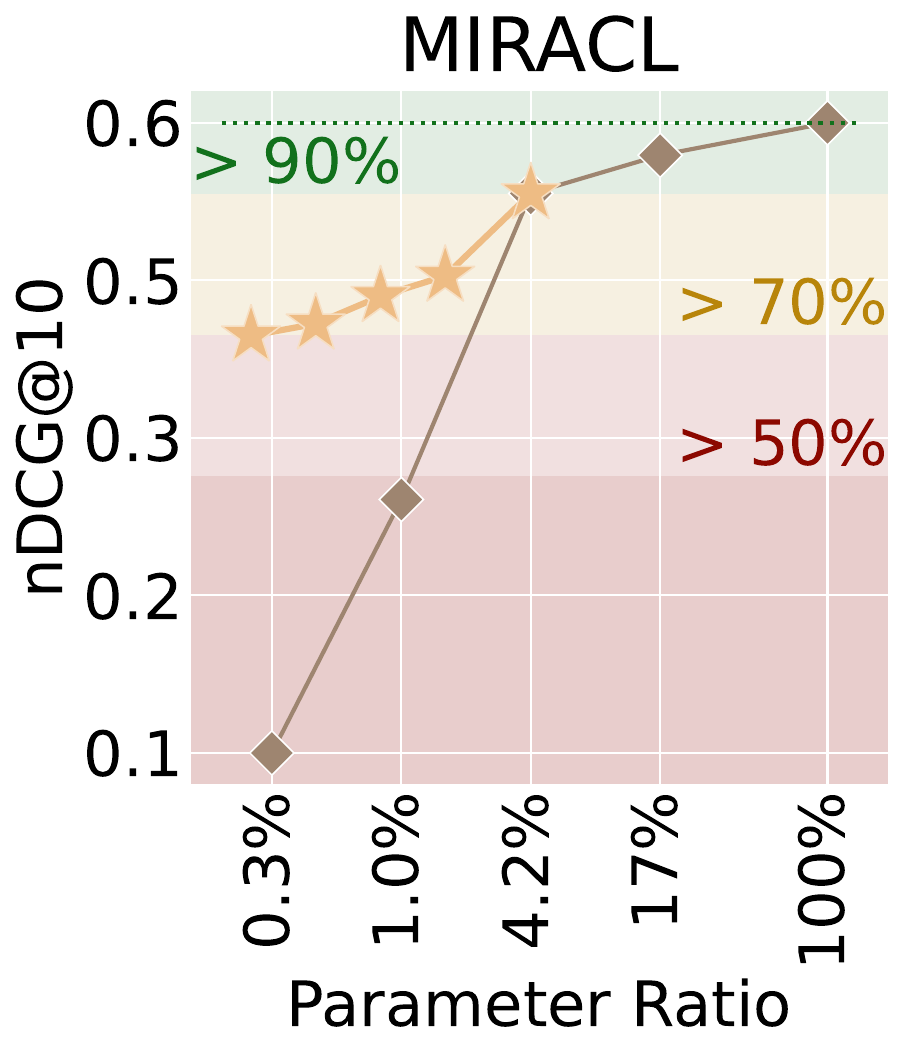}
    \end{subfigure}

    \begin{subfigure}[t]{0.54\textwidth}
        \includegraphics[width=1\columnwidth,trim={0cm 0cm 0cm 0},clip]{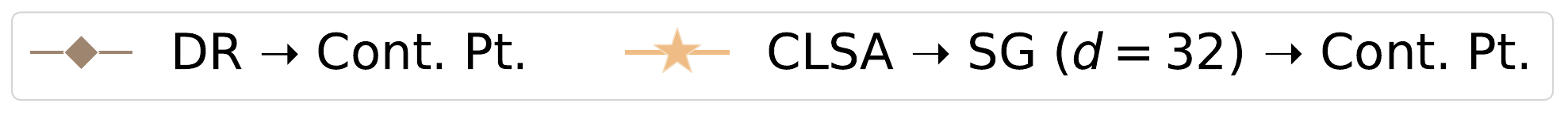}
    \end{subfigure}

\caption{
    Comparison the effectiveness of Dimension Reduction (DR) and semantic Grouping (SG) at the same level of word embedding parameters size. 
    {\bf x-axis}: the ratio of the word embedding size after DR or SG in log scale.
    The \browncolor lines:\ DR with dimension $d \in \{2, 8, 32, 128, 768\}$ and full vocabulary,
    corresponding to embedding parameters ratio of \{0.3\%, 1.0\%, 4.2\%, 17\%, 100\%\}.
    The \yellowcolor lines:\ SG with dimension $d=32$ and vocabulary grouping ratio $r_G \in \{5\%, 10\%, 20\%, 40\%\}$,
    corresponding to embedding parameters ratio of \{0.21\%, 0.42\%, 0.84\%, 1.7\%, 4.2\%\}.
    Background colors design is identical to \autoref{fig:compareRandom768}.
}
\label{fig:hidden-dim-vs-vocab}
\end{figure*}

\begin{figure*}[t]
    \centering
    \begin{subfigure}[t]{0.195\textwidth}
        \includegraphics[width=1\columnwidth,trim={0 0cm 0cm 0},clip]{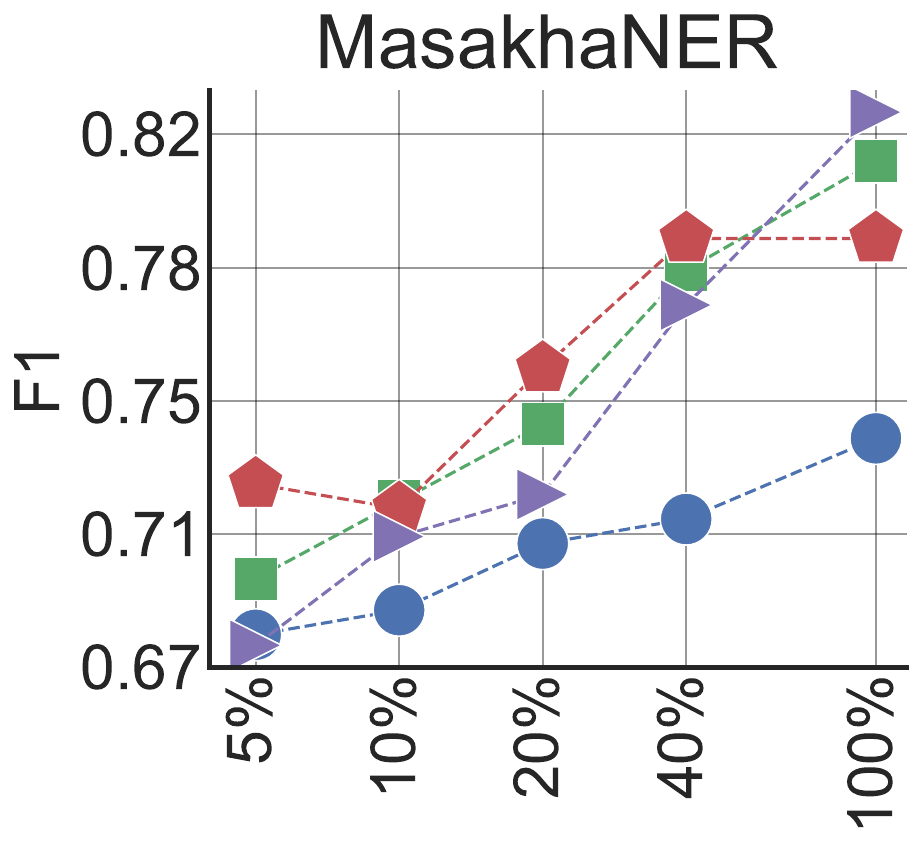}
    \end{subfigure}
    \hfill
    \begin{subfigure}[t]{0.195\textwidth}
        \includegraphics[width=1\columnwidth,trim={0 0cm 0cm 0},clip]{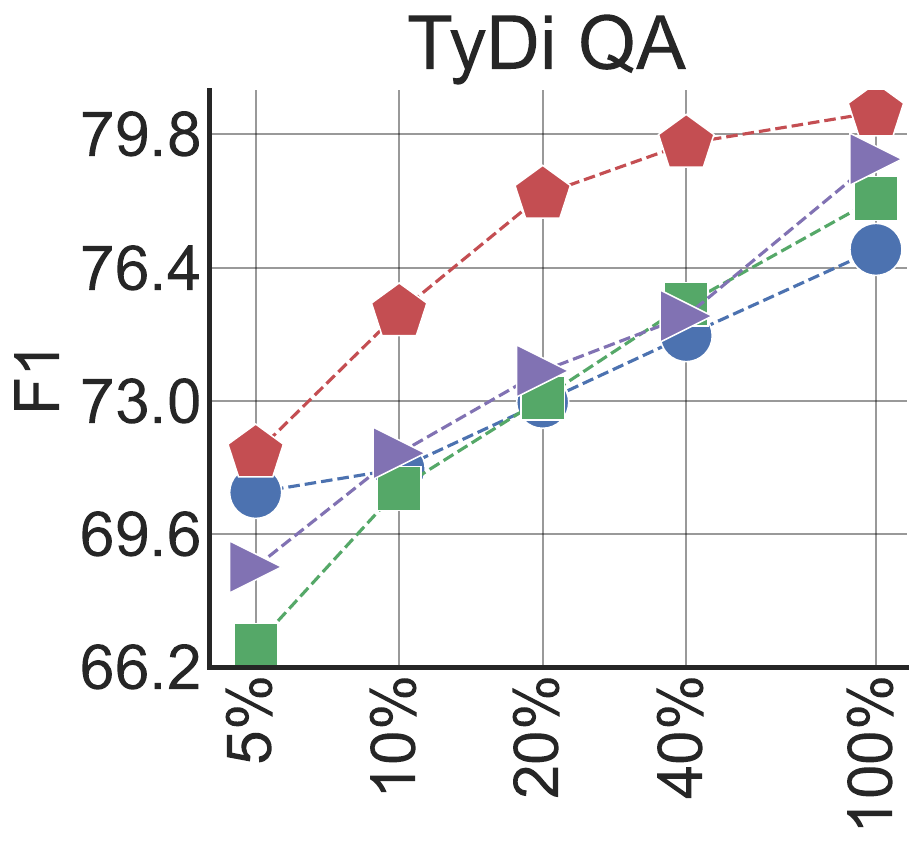}
    \end{subfigure}
    \hfill
    \begin{subfigure}[t]{0.195\textwidth}
        \includegraphics[width=1\columnwidth,trim={0 0cm 0cm 0},clip]{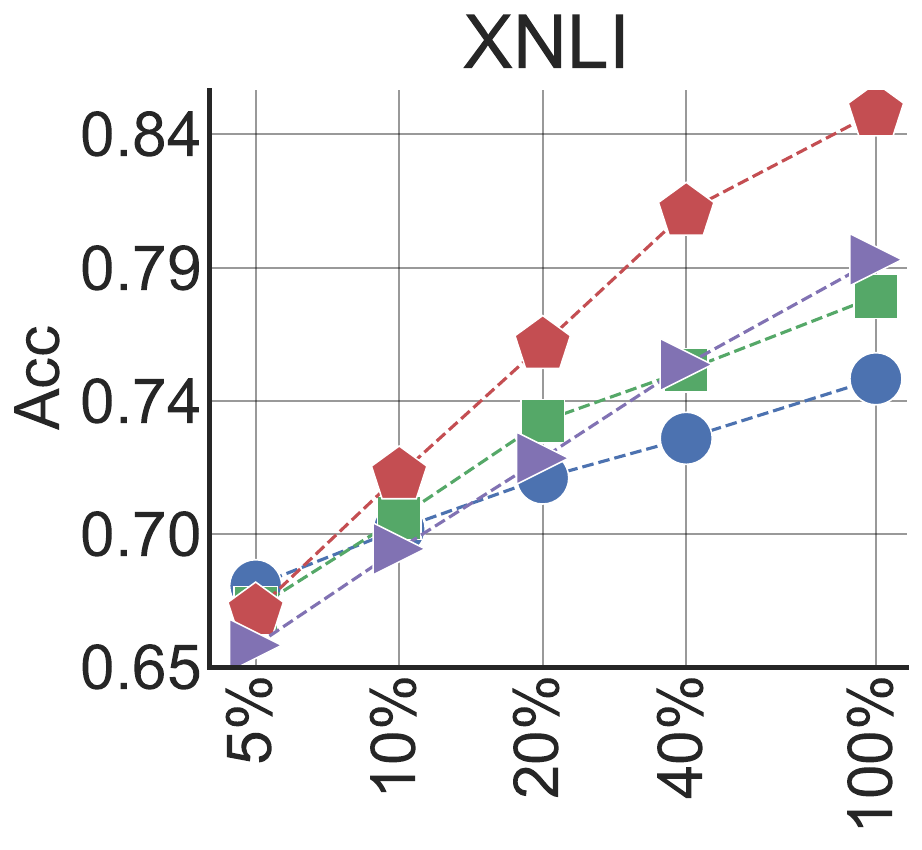}
    \end{subfigure}
    \hfill    
    \begin{subfigure}[t]{0.195\textwidth}
        \includegraphics[width=1\columnwidth,trim={0 0cm 0cm 0},clip]{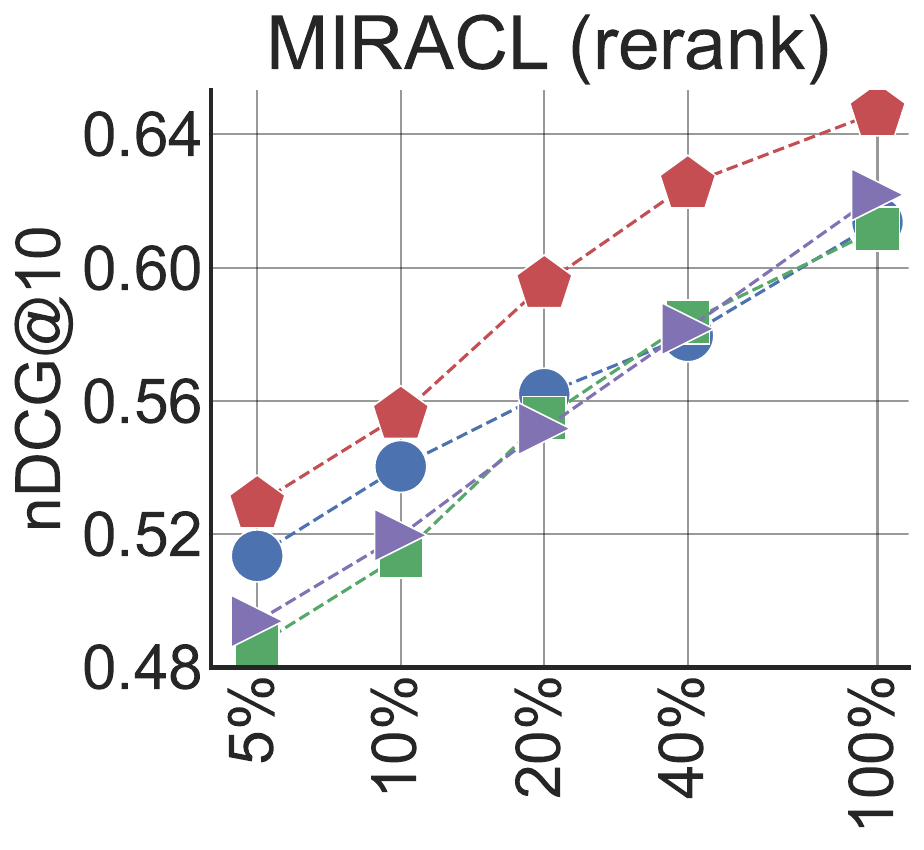}
    \end{subfigure}
    \hfill 
    \begin{subfigure}[t]{0.195\textwidth}
        \includegraphics[width=1\columnwidth,trim={0 0cm 0cm 0},clip]{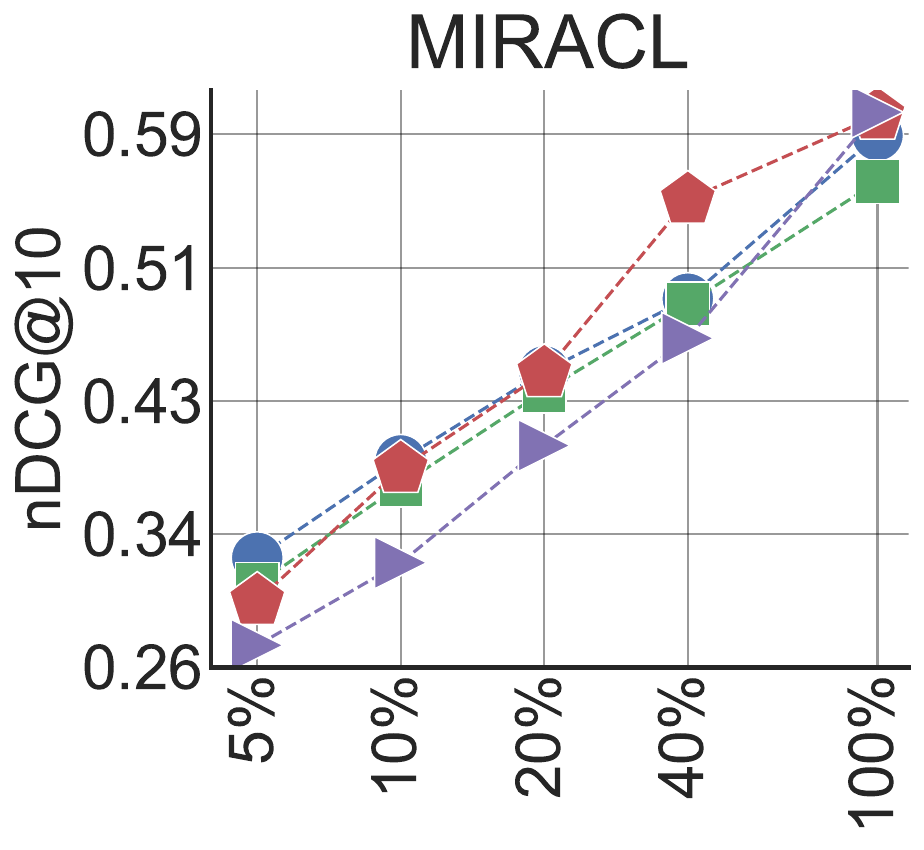}
    \end{subfigure}

    \begin{subfigure}[t]{0.7\textwidth}
        \includegraphics[width=1\columnwidth,trim={0cm 0cm 0cm 0},clip]{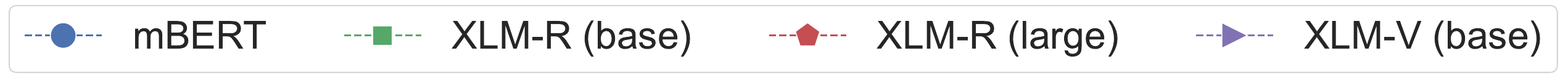}
    \end{subfigure}

\caption{
    Results of SG on multiple backbones, all with CLSA applied and no continual pretraining.
    All models show similar trend regarding semantically grouped subwords.
}
\label{fig:backbones}
\end{figure*}

\parheader{Comparison with \firstk}
\autoref{fig:compareRandom768} also shows the results of \firstk (the \redcolor lines)
described in Section~\ref{sec:method:sg}.
This is 
to compare SG with mLMs that have the same number of word embedding entries by adopting a smaller vocabulary size.
As the figure shows,
while \firstk could still share similar effectiveness with SG at $r_G = 40\%$,
removing more subwords from the vocabulary deteriorates the effectiveness drastically on all downstream tasks,
regardless of whether continual pretraining is applied.
This pair of results suggests that 
while simply reducing the number of subwords has a detrimental effect on the mLMs capacity,
which echoes previous findings~\cite{
conneau-etal-2020-unsupervised,liang-etal-2023-xlm,ali-etal-2024-tokenizer,tao2024scalinglawsvocabularylarger},
not all subwords require a unique representation.
In other words,
the effect of the number of the word embedding entries should be disentangled from the size of the vocabulary.

\parheader{Enhance the semantic similarity via post-hoc operations}
In the above results, the semantic grouping is applied on the off-the-shelf mLMs, exploiting the spatial structure of the untouched word embedding.
{\it Could the similarities of tokens under the same semantic concepts be further enhanced by post-hoc operations on the mLMs?}
We show that this direction is possible and promising,
using the cross-lingual subword alignment (CLSA) operation as an example.

\autoref{fig:768align} compares the results of applying SG on mBERT \orange{with} and \blue{without} CLSA,
where the orange lines are consistently higher than or similar to the blue ones regardless of whether the model has been continually pretrained.
On certain datasets, e.g., \xnli and \miracl, CLSA brings visible improvement at all grouping ratios,
with more significant improvements at lower grouping ratios on the other datasets.
This shows that the semantic similarity among the subwords has room to be improved by post-hoc operations.\footnote{
Note that CLSA is used as a proof-of-concept instead of a general solution:\
As mentioned in Section~\ref{sec:method:clsa},
CLSA is applied to words that are tokenized into single subword, 
which limits the coverage on the entire vocabulary intrinsically.
Thus instead of promoting the CLSA method itself,
we use its success in enhancing the semantic similarity as a proof-of-concept that further post-hoc operations on the word embedding can be beneficial and worth further exploration.
}

\subsection{Eliminating the Confounding Effect of Embedding Parameters}
\label{sec:results:dimensionReduction}

While applying the SG, 
the number of the word embedding parameters is also reduced by the same degree.
Do the changes on effectiveness truly stem from the semantic grouping,
or any form of word embedding parameters reduction have the same effect?
To address the concern,
we apply SG on word embeddings with reduced embedding dimension via the dimension reduction (DR) operation, 
to see whether the effectiveness diminishes as the embedding parameters are reduced.
We first show that SG maintains the same level of effectiveness on the reduced dimensions.
Moreover, 
while the embedding dimension has reached its limit,
SG could further push down the overall word embedding parameters to a level that could not be achieved by reducing the word embedding dimension alone.

\parheader{SG maintains the same level of effectiveness with reduced embedding dimensions}
\autoref{fig:3dim} shows the results of SG with three word embedding dimensions $d \in \{768, 128, 32\}$,
with both CLSA and continual pretraining applied.
While DR reduces the effectiveness of the mLM with full vocabulary,
the slope of the curve flattens as the dimension reduces,
indicating less relative effectiveness drop on the downstream tasks at lower embedding dimension $d$. 
Specifically,
the SG results on embedding dimension $d=32$ is on par and even outperform the results on the initial dimension $d=768$ as vocabulary size reaches 5\%.
These results show that the effect of SG is largely independent from the embedding parameters.

\parheader{SG achieves lower number of embedding parameters beyond dimension reduction} 
Apart from the above observation,
\autoref{fig:hidden-dim-vs-vocab} unites the results of SG at embedding dimension $d=32$ and DR from the perspective of the total number of word embedding parameters (x-axis).
The \browncolor lines show the DR results of reducing the embedding dimension $d$ from $768$ to $\{2, 8, 32, 128\}$ with the vocabulary untouched,
yielding to embedding parameters ratio of \{0.3\%, 1.0\%, 4.2\%, 17\%\}.
With DR, the effectiveness on all downstream tasks is largely preserved until dimension $d=32$ 
and drop sharply once the dimension falls beneath it,
indicating that saving the embedding parameters via dimension reduction has reach its limit.

On the other hand,
the \yellowcolor lines show the models 
starting from word embedding with dimension $d=32$ (thus 4.2\% embedding parameters) and then applied SG with grouping ratio $r_G$ from 40\% to 5\%,\footnote{~Identical to the yellow lines in \autoref{fig:3dim}.
}
yielding to embedding parameters ratio from $4.2\%$ to 0.21\%.
The effectiveness differences between the two lines are clear:\
while further reducing the embedding parameters by similar scale, 
SG could largely preserve the downstream effectiveness whereas DR fails miserably.
This strongly informs that SG is complementary to DR on their effect towards the word embedding parameters,
and that it may provide a new perspective on understanding the necessary parameters in the word embeddings.

\subsection{Backbones}
\label{sec:results:backbones}
Experiments above are all based on mBERT~\cite{devlin-etal-2019-bert},
how would the findings generalize to mLMs with different tokenization algorithms, vocabulary size, model size, and pre-training corpora?
We select three additional mLMs to address the above concern:\
XLM-R (base), XLM-R (large)~\cite{conneau-etal-2020-unsupervised}, and XLM-V (base)~\cite{liang-etal-2023-xlm},
which all deploy ULM~\cite{kudo-richardson-2018-sentencepiece} to construct the vocabulary, and pretrained on CC100, 
but differ from each other in terms of model sizes, tokenization pre-processing, vocabulary allocation, and vocabulary size.

Results of SG are shown in \autoref{fig:backbones},
where background colors are omitted as different backbones do not share the same oracle results. 
Instead, we compare the results of the other backbones with mBERT (the \bluecolor lines) to see whether they follow similar trend as more subwords are semantically grouped. 
Overall, the slopes of the four curves are similar on all five benchmarks, 
indicating that the findings are likely to generalize over different multilingual LMs with various tokenization processes and model sizes.

\subsection{Semantic Similarities Categories among the Grouped Subwords}
\label{sec:analysis}
\begin{figure}[t]
    \centering
    \includegraphics[width=\columnwidth,clip,trim={0cm 0 0cm 0}]{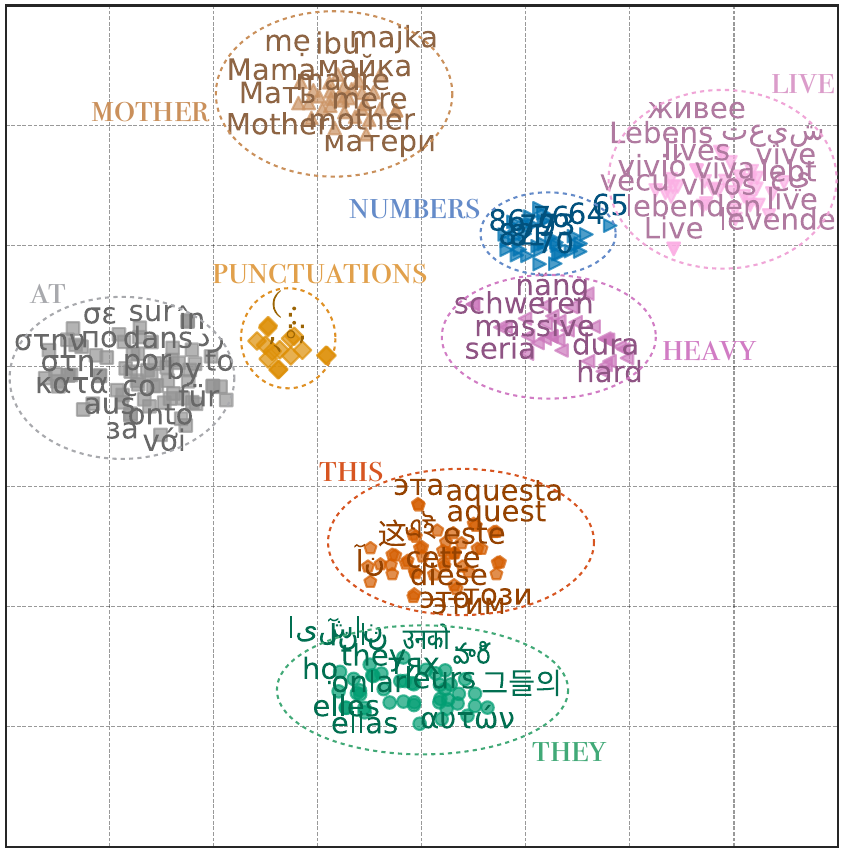}

    \caption{
        Eight semantic tokens formed by grouped subwords on mBERT word embedding with dimension $d=768$ and CLSA applied.
        The labels outside the circles are either the keyword of the cluster,
        or ``{\small \texttt{NUMBERS}}''/ ``{\small \texttt{PUNCTUATIONS}}'' if the cluster is a collection of numbers or punctuations.
    }
    \vspace{-0.8em}
    \label{fig:inspectionCluster}
\end{figure}

This section provides insights into the semantic similarities among grouped subwords.
\autoref{fig:inspectionCluster} illustrates examples of eight semantic groups, 
which are based on the model with embedding dimension $d=768$ and the application of CLSA.\footnote{~Note that only parts of the subwords are displayed per group for space limit.
For the full list of subwords, please refer to Ap.~\ref{ap:inspection} and \autoref{fig:ap:inspection}.
}

Among the eight displayed groups examples displayed,
six of them are selected from the Swadesh list~\cite{swadesh1952lexico},
a widely used compilation of basic concepts across languages.
The keywords are selected to cover various parts of speech, including pronouns, nouns, adjectives, prepositions, and verbs.
The remaining two groups are labeled as ``{\small \texttt{NUMBERS}}'' and ``{\small \texttt{PUNCTUATIONS}}'',
which exhibit strong clustering upon manual inspection.
Overall, we identify several patterns of semantic similarities among the grouped subwords:\
\begin{enumerate}[itemsep=-3pt, topsep=8pt, leftmargin=14pt]
    \item Semantically identical or similar words across languages:\
    {\it mother vs.\ \protect\raisebox{-3pt}{\includegraphics[width=5mm]{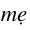}}, they vs.\ \protect\raisebox{-3pt}{\includegraphics[width=9mm]{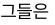}}.\footnote{~{\protect\raisebox{-3pt}{\includegraphics[width=5mm]{images/fonts/mother_vie.pdf}}: ``mother'' in Vietnamese; {\it Cette}: ``this'' in French; \protect\raisebox{-3pt}{\includegraphics[width=9mm]{images/fonts/they_ko.pdf}}: ``they'' in Korean.}}}
    \item Semantically related words from the same language:\ {\it at vs.\ on, heavy vs.\ hard}.
    \item Semantically related words across languages:\ {\it heavy vs.\ intenso, who vs.\ \protect\raisebox{-3pt}{\includegraphics[width=6mm]{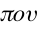}}}.\footnote{~{\it intenso}: ``intense'' in Spanish; \protect\raisebox{-3pt}{\includegraphics[width=6mm]{images/fonts/who_greek.pdf}}: ``where'' in Greek.
    }
    \item Numbers in similar range:\ {\it the cluster numbers ranging from 60 to 100.}
    \item Punctuations: {\it ., ;!}
\end{enumerate}

\smallskip \noindent
We also notice that some words are not grouped based on desired semantics:\
The current methods are limited in the polysemous situation and controlling the desired semantic per group:\ 
For example,
in the group of ``I'',
while the desired semantic is the first person singular pronouns in different languages,
the group mainly includes the single letter such as ``A'', ``B'', ``C'' on mLMs with $d=768$, 
as shown by \autoref{fig:ap:inspection} in Ap.~\ref{ap:inspection},
suggesting that future work is necessary to better handle such cases.

\begin{figure}[t]
    \centering
        \begin{subfigure}[t]{0.49\columnwidth}
            \includegraphics[width=1\columnwidth,trim={0.3cm 0cm 0cm 0},clip]{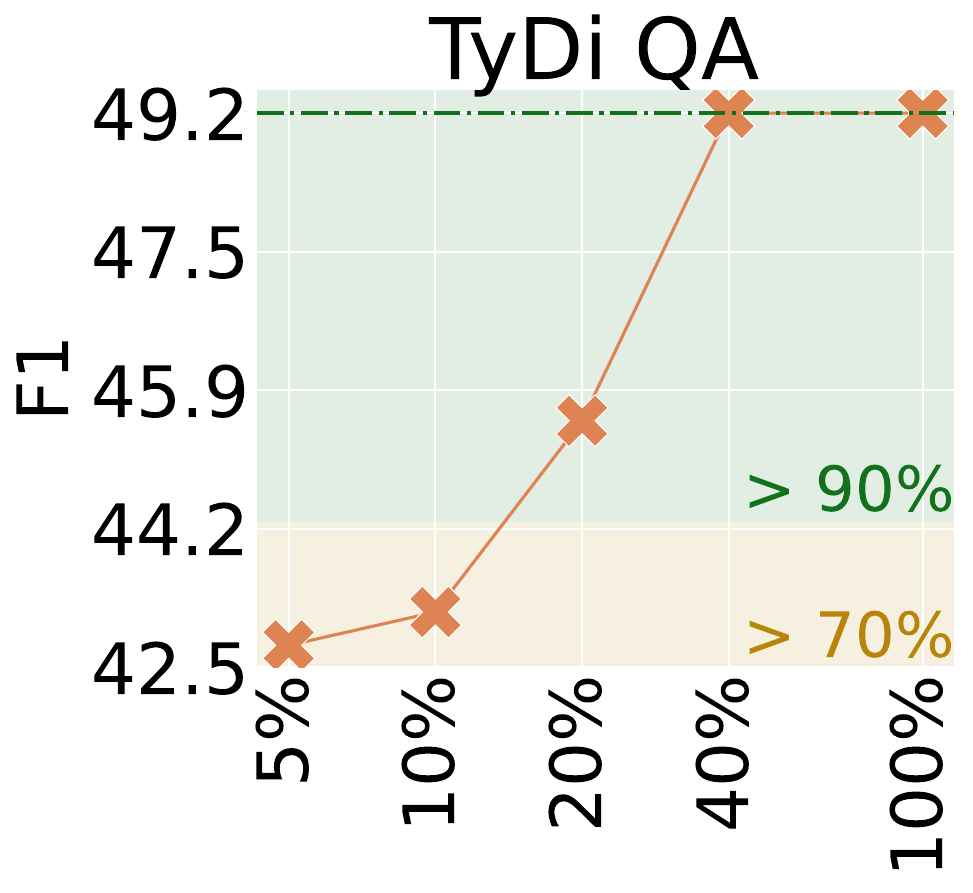}
        \end{subfigure}
        \hfill
        \begin{subfigure}[t]{0.49\columnwidth}
            \includegraphics[width=1\columnwidth,trim={0.3cm 0cm 0cm 0},clip]{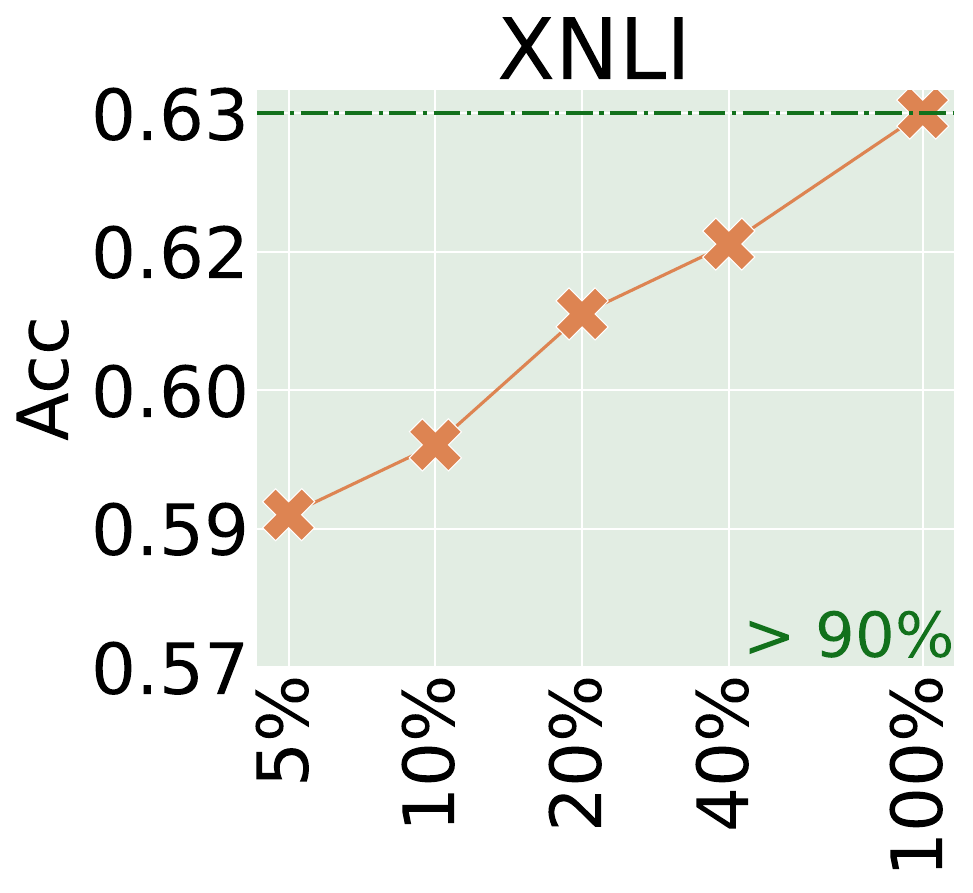}
        \end{subfigure}

        \begin{subfigure}[t]{0.49\columnwidth}
            \includegraphics[width=1\columnwidth,trim={0.3cm 0cm 0cm 0},clip,right]{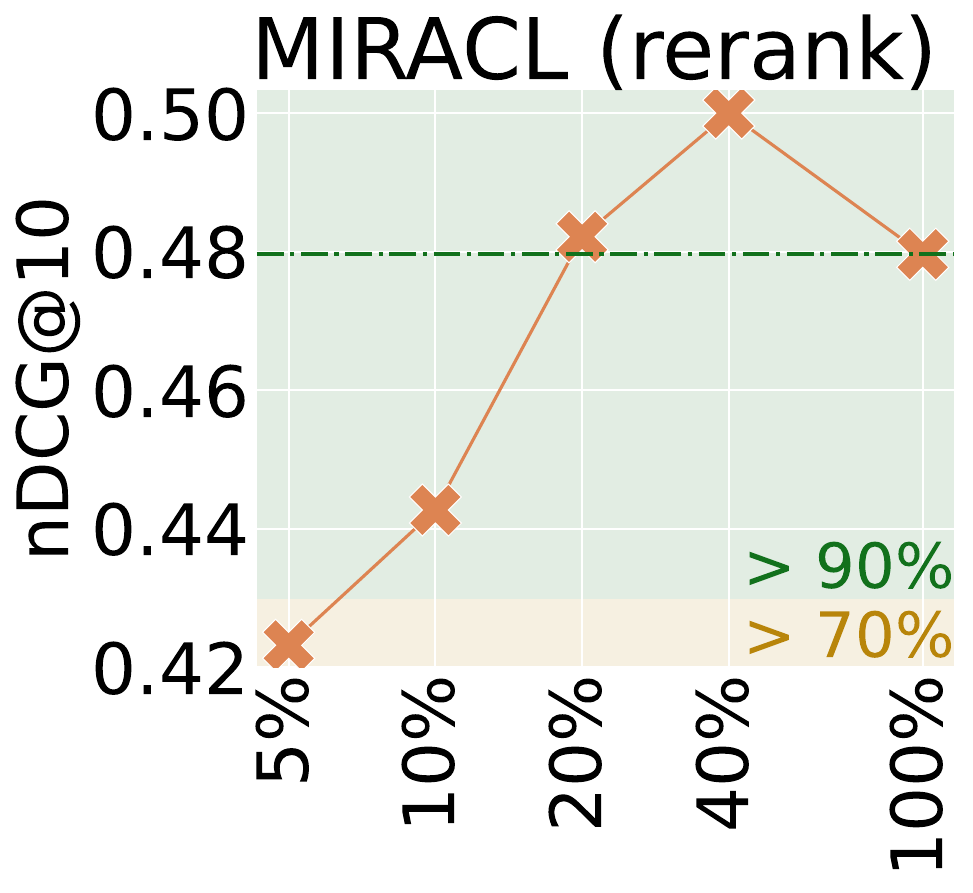}
        \end{subfigure}
        \hfill 
        \begin{subfigure}[t]{0.49\columnwidth}
            \includegraphics[width=1\columnwidth,trim={0.3cm 0cm 0cm 0},clip,left]{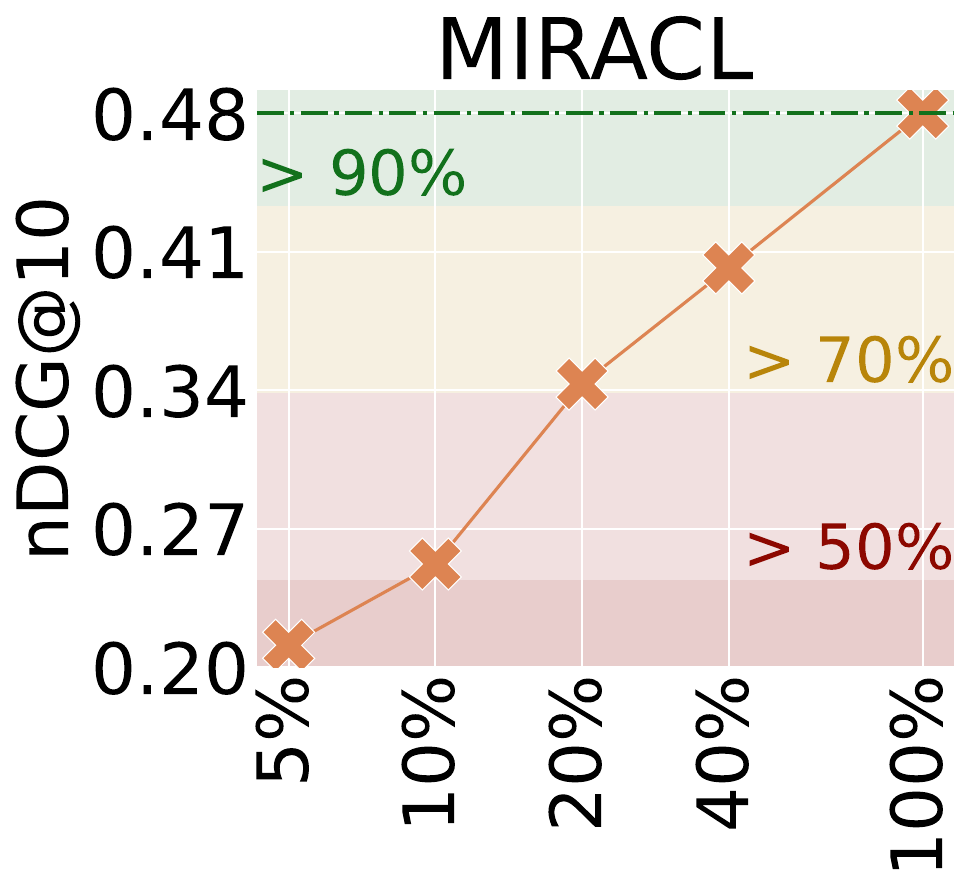}
        \end{subfigure}

    \caption{
        Results when transferring from English to other languages on four benchmarks,
        with CLSA and continual pretraining applied.
        \masakhaner is skipped as English training data is not available.
        Background colors design is identical to \autoref{fig:compareRandom768}.
        The dashed green line denotes the zero-shot performance without semantic grouping.
    }
    \label{fig:zero-shot}
\end{figure}

\subsection{Cross-lingual Transfer}
\label{sec:results:zero-shot}

Lastly,
we investigate how the semantic grouping affects the cross-lingual transferability from English to other languages.
We evaluate on four of the above tasks,
where \masakhaner is skipped in this section as English training data is not available.

Surprisingly,
on two classification tasks:\ \tydiqa and \miracl (rerank),
the zero-shot results at grouping ratio $r_G \in \{40\%, 20\%\}$ are on par or even better than the oracle results, 
where subwords and embeddings are not semantically grouped. 
That is, fine-grained semantics show similar cross-lingual transferability with word-based understanding in some circumstances,
suggesting that they may serve as the anchors for cross-lingual transferring in the current language models.

On the other hand,
cross-lingual transfer on embedding tasks is more challenging especially with small group ratio, i.e., coarse-grained semantic tokens.
While the zero-shot results on \miracl at $r_G \in \{20\%, 40\%\}$ 
remains in the reasonable range --- achieving more than $70\%$ of the relative zero-shot effectiveness ---
the relative zero-shot results falls under 50\% when $r_G$ further drops.
This mirrors the in-domain results from previous sections that embedding tasks require more fine-grained semantic information to form effective sentence or paragraph representations.

\section{Related Works}

\parheader{Semantic Similarities in Language Models}
The semantics latent knowledge in word embeddings have been leveraged to adapt the pretrained multilingual LMs to unseen languages or scripts~\citep{pfeiffer-etal-2021-unks,wang-etal-2022-expanding,liu-etal-2023-crosslingual-transfer}.
While it has been shown that the shared semantics assist the language transfer, 
it is unclear how much the semantics alone could achieve compared to the full-fledged models.
Another line of works focuses on the isomorphic representations of {\it contextualized} word embeddings. 
\citet{li2024vision} find that vision and language models shows isomorphic representations,
and \citet{peng-sogaard-2024-concept} reveal similar conclusion on the contextualized representations of multilingual large language models.
Two additional works also study the contextualized concept encoded in the language models:\
\citet{sajjad-etal-2022-analyzing} analyze how latent concepts are encoded in representations and how they align with human-defined concepts.
\citet{shani-etal-2023-towards} analyze how large language models encode the \textsc{TypeOf} relations of concepts, and propose a model-agnostic, proof-of-concept method to shift the model to a concept-level understanding.

\parheader{Bilingual Lexicon Induction and Word Alignment}
The Cross-lingual Subword Alignment (CLSA) operation is related to the task of {\it Bilingual Lexicon Induction (BLI)} and {\it Word Alignment}.
BLI aims to induce the equivalent translation in language $L_2$ given a word in language $L_1$~\cite{artetxe-etal-2016-learning,conneau2017word,wang2020crosslingual,shi-etal-2021-bilingual}.
CLSA is similar to BLI in terms of focusing on uncontextualized cross-lingual words pairs,
but different in that it aims to align the word pairs in the embedding space and targets on not only bilingual but also multilingual words.
Word alignment aims to find bilingual word pairs in parallel sentences~\cite{Cao2020Multilingual,jalili-sabet-etal-2020-simalign}.
While names are similar, the goals are different:\
In addition to above differences with BLI, 
CLSA focuses on uncontextualized cross-lingual word pairs and does not involve parallel sentences.

\parheader{Word Sense Clustering}
The Semantic Grouping (SG) operation is related to the task of word sense clustering.
Although both share similar objectives, word sense clustering primarily relies on corpus statistics~\cite{snow-etal-2007-learning}, whereas the SG operation is based on the similarity of word embeddings.

\parheader{Parameter Redundancy and Model Compression}
\citet{dalvi-etal-2020-analyzing} study the layer and neuron redundancy on BERT and XLNet,
and many works propose to compress the overall model size via pruning~\cite{gordon-etal-2020-compressing,ashkboos2024slicegpt,yang2024laco},
knowledge distillation~\cite{turc2019well,sanh2019distilbertad},
and quantization~\cite{shen_dong_ye_ma_yao_gholami_mahoney_keutzer_2020}.
From the perspective of model compression,
our work provides a new view on the word embeddings redundancy from the perspective of shared semantics among subwords.

\section{Conclusion}

Inspired by how human understand text based on semantic concepts rather than superficial forms,
this work studies the degree to which current multilingual language models understand based on subword-level semantic concepts.
We find that the general shared semantics could get the models a long way in understanding languages and making predictions, especially for classification tasks.
Additional experiments show that the observations generalize across mLMs with different tokenization algorithm, vocabulary size, model size, and pre-training corpora.
Inspections on the grouped subwords suggest that they exhibit multiple patterns of semantic similarities,
including synonyms and word translations in many languages.

Not only the subword-level semantics is prominent in in-domain language understanding,
in some cases, it also serves as anchors of cross-lingual transfer 
and thus potentially a promising direction of bridging the understanding of different languages.
We hope that this work sheds light on understanding the multilingual vocabulary and word embeddings from the semantic perspective,
and spurs further research on shared semantic information at the subword level across languages.

\section*{Limitations}

\parheader{The Scope of Semantics}
This work only discusses the application scenario where pragmatics are not heavily involved. 
Other situations such as poetry, humour, sentiment analysis intuitively would require not only the semantic meanings, but also exquisite understanding of the words nuances,
yet out of the scope of this work.
Similarly, the work is probably not applicable to figurative language such as metaphor, irony, etc. 

\newpage
\parheader{Word-level Semantics Only}
One of the major limitation of this work is not consider the phrase-level semantics in the study.

\parheader{Encoder-only Models}
As a natural limitation of the semantical grouping method itself,
it is not straightforward to extend the method to decoder-only models since it forbids predicting explicit subword at each decoding step.
Thus only the encoder-only models and tasks are evaluated in this work.
Further design and exploration would be required to apply the method to decoder-only models.

\parheader{Results on Embedding Tasks}
Results show that the embedding tasks are more sensitive to the semantical grouping compared to the classification tasks.
More questions could be raised from the phenomenon:\
does it make embedding task a better evaluation metrics for the semantical grouping, 
or that simply the embedding task require more fine-grained understanding of the subword information?
How much could embeddings benefit if the semantical grouping algorithm could be improved?
We believe that these are important questions to further understand effectiveness and {\it limitation} of this direction.
Limited by the paper capacity, we leave them for future exploration.

\section*{Acknowledgement}
We greatly thank Jiarui Xu, Xinyu Shi, Xueguang Ma, Raphael Tang, Sanket Vaibhav Mehta, and the anonymous reviewers for their valuable discussions and feedback.

\bibliography{anthology,custom,camera}
\bibliographystyle{acl_natbib}

\clearpage

\appendix

\section{Training and Evaluation Configurations}
\label{ap:config}
\parheader{Continual Pretraining}
All continual pretraining in this work share the same hyperparameters.
Language models are trained on the MLM objective for 25,000 steps, with a batch size 1024 and a learning rate of $1e-4$. We randomly masked 15\% of the tokens.
The hyperparameters were chosen following the initial pretraining configurations of \citet{devlin-etal-2019-bert}.

\parheader{Downstream Task}
For each task, we use all the available training data provided by each dataset.
The training configurations are provided in \autoref{tab:ap:downstream-config}.
\begin{table}[t]
    \resizebox{\columnwidth}{!}{
    \begin{tabular}{lrrrrr}
    \toprule
    \multirow{2}{*}{} & 
        \multicolumn{1}{l}{\multirow{2}{*}{\bf mNER}} & 
        \multicolumn{1}{l}{\multirow{2}{*}{\textbf{\tydiqa}}} & 
        \multicolumn{1}{l}{\multirow{2}{*}{\textbf{\xnli}}} & 
        \multicolumn{2}{c}{\textbf{MIRACL}} \\
     & \multicolumn{1}{l}{} & \multicolumn{1}{l}{} & \multicolumn{1}{l}{} & \multicolumn{1}{c}{\textbf{rerank}} & \multicolumn{1}{c}{\textbf{retrieval}} \\
    \midrule
    epochs & 50 & 3 & 2 & 5 & 40 \\
    warmup ratio & \multicolumn{1}{r}{--} & \multicolumn{1}{r}{--} & \multicolumn{1}{r}{--} & 0.1 & 0.1 \\
    batch size & 64 & 128 & 64 & 32 & 256 \\
    learning rate & 5e-05 & 5e-05 & 5e-06 & 5e-06 & 1e-05 \\
    adam $\beta_1$ & 0.9 & 0.9 & 0.9 & 0.9 & 0.9 \\
    adam $beta\_2$ & 0.99 & 0.99 & 0.99 & 0.99 & 0.99 \\
    \bottomrule
    \end{tabular}
    }
\caption{
    Downstream task traning configurations.
    {\bf mNER:\ \masakhaner}
}
\label{tab:ap:downstream-config}
\end{table}



\section{Alignment Datasets}
\label{ap:aligndata}
\autoref{fig:ap:alignment-source} compares the impact of different word alignment datasets on the downstream tasks.
All experiments are followed by $r_G=5\%$ semantic grouping and continual pretraining.
On all downstream datasets, we compare the results of using four groups of alignment data:\
\begin{enumerate}[noitemsep,topsep=0pt]
    \item MUSE
    \item MUSE and Round-Trip 
    \item MUSE and PanLex
    \item MUSE, PanLex, Colex, and Concepticon
\end{enumerate}
where Round-Trip are the pair of word that are the nearest neighbors to each other in the embedding space,
serving as a regularization in the CLSA procedure. 
We found that scenario~4 gives the best overall results, and thus use it as our default configuration.

\begin{figure}[t]
    \vspace{-1em}
    \centering
    \includegraphics[width=\columnwidth,clip,trim={0 0 0 0}]{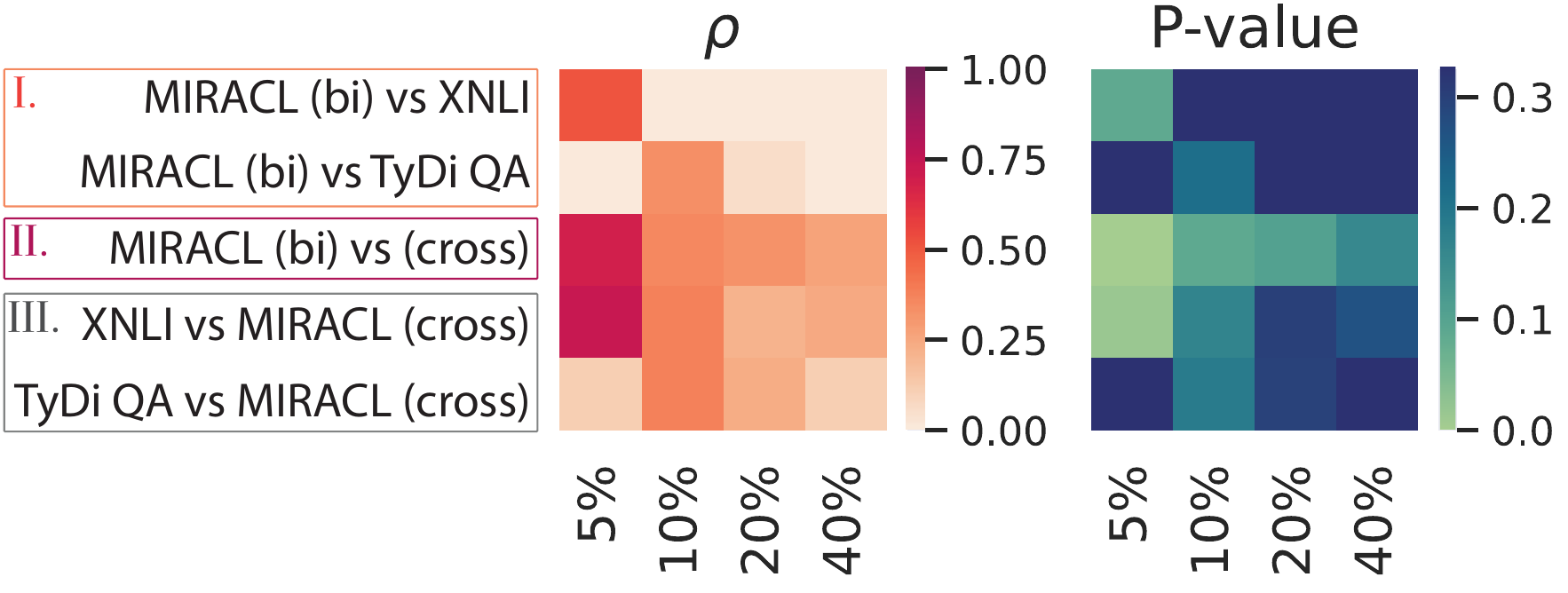}
    \caption{
        Pearson correlation between the relative performance drop per language between a pair of benchmarks. 
        Each row indicates a pair of benchmarks, where (bi) refers to retrieval (which uses bi-encoder) and (cross) refers to reranking (which uses cross-encoder).
        Each column indicates a semantic grouping rate $r_G$.
        {\it Left}: Pearson correlation coefficient $\rho$; 
        {\it Right}: corresponding one-tail p-values. 
    }
    \label{fig:single-lang:correlation}
\end{figure}
\begin{figure*}[!h]
    \centering
    \begin{subfigure}[t]{\columnwidth}
        \centering
        \includegraphics[width=0.95\columnwidth,clip,trim=7mm 0 0 0]{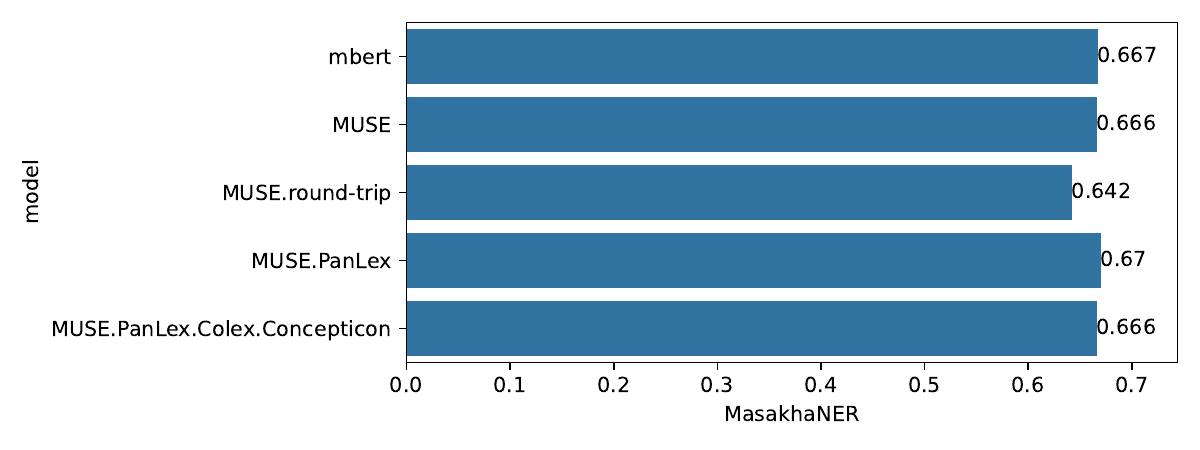}
    \end{subfigure}
    \hfill
    \begin{subfigure}[t]{\columnwidth}
        \centering
        \includegraphics[width=0.95\columnwidth,clip,trim=7mm 0 0 0]{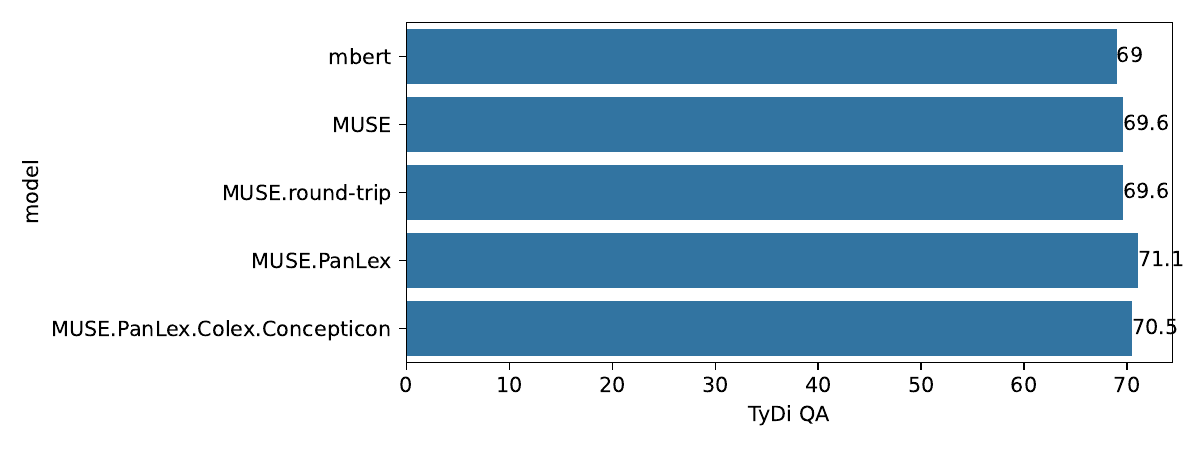}
    \end{subfigure}
   
    \begin{subfigure}[t]{\columnwidth}
        \centering
        \includegraphics[width=0.95\columnwidth,clip,trim=7mm 0 0 0]{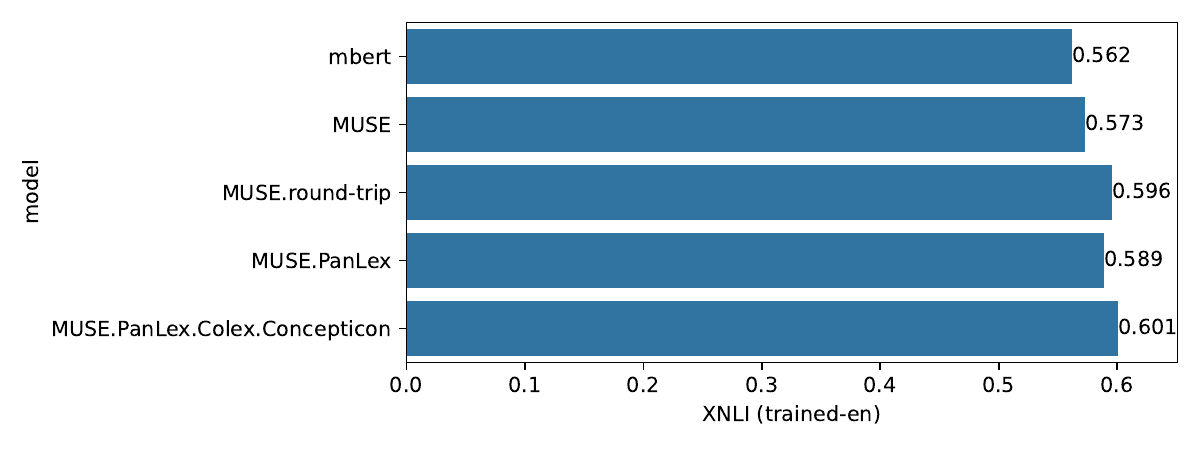}
    \end{subfigure}
    \hfill
    \begin{subfigure}[t]{\columnwidth}
        \centering
        \includegraphics[width=0.95\columnwidth,clip,trim=7mm 0 0 0]{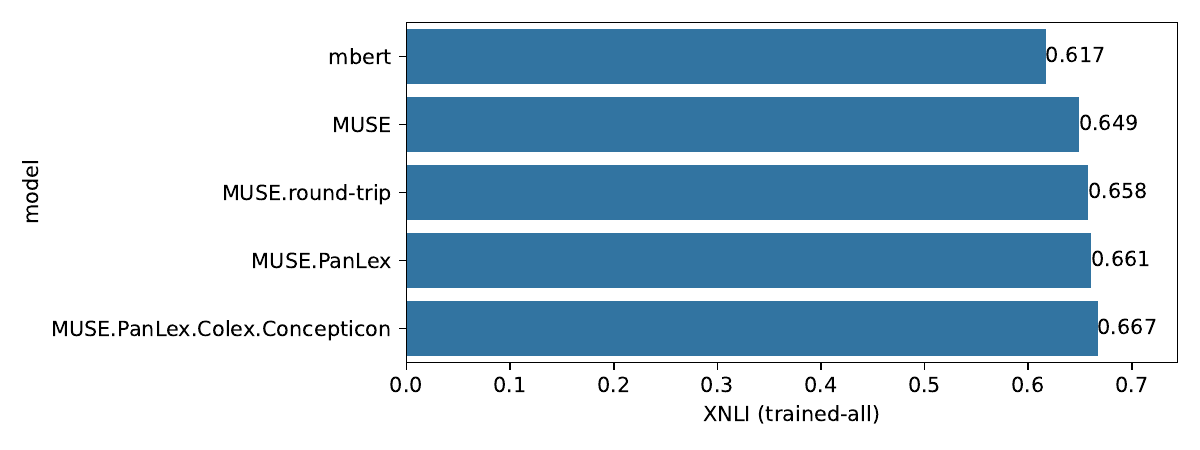}
    \end{subfigure}

    \begin{subfigure}[t]{\columnwidth}
        \centering
        \includegraphics[width=0.95\columnwidth,clip,trim=7mm 0 0 0]{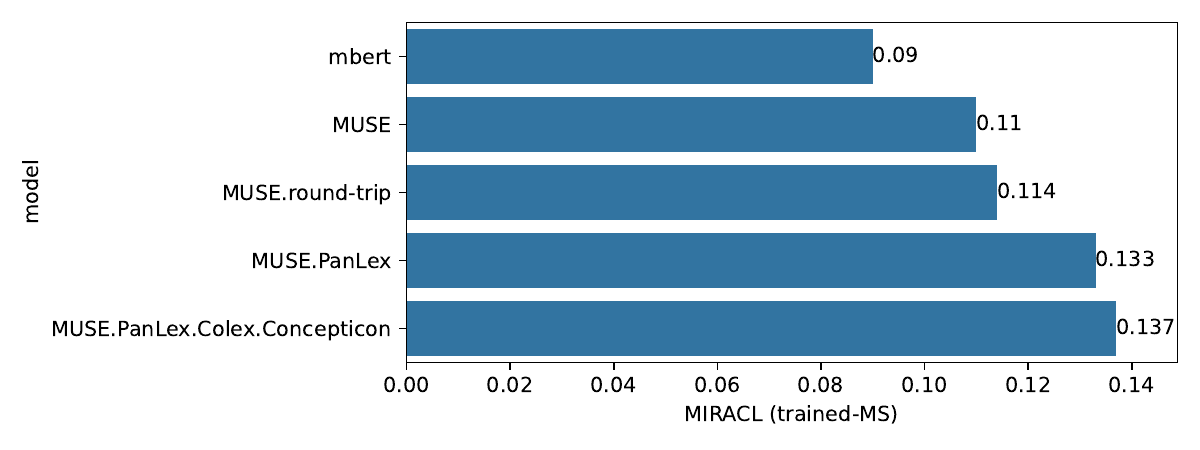}
    \end{subfigure}
    \hfill    
    \begin{subfigure}[t]{\columnwidth}
        \centering
        \includegraphics[width=0.95\columnwidth,clip,trim=7mm 0 0 0]{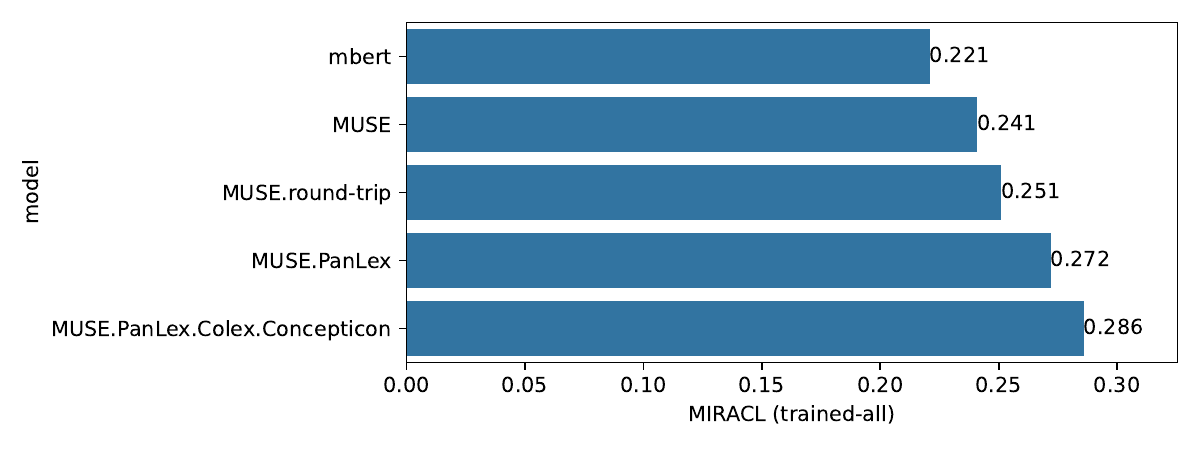}
    \end{subfigure}
    
\caption{Comparison of alignment source dataset on downstream tasks. 
Each bar represents one alignment dataset(s), where ``mbert'' is the baseline result when there is no alignment applied.
The x-axis is the average score on all languages per task.
}
\label{fig:ap:alignment-source}
\end{figure*}

\section{Impact on Individual Languages}
\label{ap:analysis:individual-lang}

This section explores whether the languages are consistently affected across tasks by the semantic grouping.
To this end,
we compare the effectiveness drop on the overlapping languages of each pair of benchmarks,
and compute their Pearson correlation coefficient. 
Five pairs of benchmarks are selected for analysis, which fall under 3 groups:\
%
\begin{enumerate}[noitemsep,topsep=0pt,leftmargin=7mm]
    \item[I.] {\bf different task types and data sources}:
        \begin{enumerate}
            \item \miracl (retrieval) vs.\ \xnli
            \item \miracl (rerank) vs.\ \tydiqa
        \end{enumerate}
    \item[II.] {\bf different tasks types, same data source}:\ \\
        \miracl (retrieval) vs.\ \miracl (rerank)
    \item[III.] {\bf same task type, different data sources}:\ 
        \begin{enumerate}
            \item \miracl (rerank) vs.\ \xnli
            \item \miracl (rerank) vs.\ \tydiqa
        \end{enumerate}
\end{enumerate}
The benchmark selection is mainly under the consideration of the number of overlapping languages:\
\masakhaner have no overlapping languages with the other datasets,
and \tydiqa and \xnli only have 3 overlapping languages.

\smallskip \noindent
\autoref{fig:single-lang:correlation} shows Pearson correlation coefficient $\rho$ (left) and the corresponding p-values (right) in two heatmaps.
In two heatmaps, higher saturation indicates higher $\rho$ or smaller p-values, respectively,
{which together indicates stronger correlations in higher confidence.}
Each row corresponds to a pair of benchmarks, and each column corresponds to a semantic grouping ratio $r_G$.
The three blocks in the figure corresponds to the three groups defined above from top to bottom.
%

Overall, we observe consistent trend on the two heatmaps,
where the top-2 rows are smaller in the coefficient $\rho$ (lighter color in the left heatmap) and larger in p-values (darker color in the right heatmap),
and that the bottom-3 rows are larger in the coefficient $\rho$ and smaller in p-values.
This indicates that 
languages are affected similarly by the semantic grouping when the benchmarks pair share either the same data source (group 2) or the task type (group 3).
In contrast, the impact on the languages is less consistent across benchmarks that shares neither the task type nor the data source (group 1).

\begin{table}[t]
    \resizebox{\columnwidth}{!}{
        \begin{tabular}{crrrr}
        \toprule
        $d$ & $r_G$ & \multicolumn{1}{l}{\textbf{\# Para (M)}} & \textbf{GPU Mem (G)} & \textbf{steps / sec} \\
        \midrule
        \multicolumn{1}{r}{768} & 5.0\% & \cellcolor[HTML]{EBEFDB}90 & \cellcolor[HTML]{D9EAD3}19.1 & \cellcolor[HTML]{DCEBD2}5.1 \\
        & 10.0\% & \cellcolor[HTML]{F7F1D2}95 & \cellcolor[HTML]{F1EFCF}39.8 & \cellcolor[HTML]{FDF2CC}2.6 \\
        & 20.0\% & \cellcolor[HTML]{FEF1CD}104 & \cellcolor[HTML]{FEF1CD}51.5 & \cellcolor[HTML]{FADDCC}2.1 \\
        & 40.0\% & \cellcolor[HTML]{FAEDD1}123 & \cellcolor[HTML]{F4CCCC}74.9 & \cellcolor[HTML]{FADECC}2.1 \\
        & 100.0\% & \cellcolor[HTML]{EFDFDF}178 & \cellcolor[HTML]{F5CDCC}74.6 & \cellcolor[HTML]{F6CECC}1.9 \\
        \midrule
        \multicolumn{1}{r}{128} & 5.0\% & \cellcolor[HTML]{E1EEE2}86 & \cellcolor[HTML]{D9EAD3}19.1 & \cellcolor[HTML]{DAEBD2}5.2 \\
        & 100.0\% & \cellcolor[HTML]{FFF2CC}101 & \cellcolor[HTML]{F5CDCC}74.6 & \cellcolor[HTML]{F6CCCC}1.9 \\
        \midrule
        \multicolumn{1}{r}{32} & 5.0\% & \cellcolor[HTML]{E1EEE2}86 & \cellcolor[HTML]{D9EAD3}19.1 & \cellcolor[HTML]{D9EAD3}5.3 \\
        & 100.0\% & \cellcolor[HTML]{E8EFDD}89 & \cellcolor[HTML]{F5CDCC}74.6 & \cellcolor[HTML]{F6CCCC}1.9 \\

        \bottomrule
        \end{tabular}
    }
    \caption{
        Efficiency statistics of mBERT under SG and DR,
        collected on a 80G A100 GPU during pretraining with batch size 128 per device.
        (\greencolor: best; \redcolor: worse; \yellowcolor: neutral)
    }
    \label{tab:efficiency}
\end{table}

\section{Discussion on Memory and Efficiency}
\label{ap:discussion-memory-efficiency}
This section discusses the effect of semantic grouping on the model size, and its memory usage and training speed during the continual pretraining.
\autoref{tab:efficiency} shows above statistics of mBERT with $r_G \in \{5\%, 10\%, 20\%, 40\%, 100\%\}$ with word embedding dimension $d=768$, and $r_G \in \{5\%, 100\%\}$ with $d \in \{128, 32\}$.\footnote{
    Statistics of LMs with $r_G \in \{10\%, 20\%, 40\%\}$ and $d \in \{128, 32\}$ are similar to $d=768$, thus skipped for simplicity.
}

\parheader{Model Size}
The model size is affected linearly with the vocabulary size or the word embedding dimension.
As the word embedding initially takes over half of the total model parameters in mBERT,
grouping the subwords to 5\% of the vocabulary size brings visible savings on the overall model size, from 178M to 90M.

\parheader{Memory}
We found that memory usage during the pretraining could be largely saved via reduced vocabulary size, but not the word embedding dimension.
We explain it by that the memory usage during the pretraining is bottlenecked by the activations, especially the final token-level logit matrix,
whose size is solely determined by the vocabulary size but not the word embedding dimension.
As a result, the memory savings from compact vocabulary is prominent, 
from 74.6G to 19.1G when reducing the vocabulary size from 100\% to 5\%,
while saving the word embedding dimension barely changes the memory usage.

\parheader{Training Speed}
The trend of training speed is similar to the memory usage,
where saving the word embedding dimension has negligible effect on the training speed
while saving the vocabulary size has significant impact, from 1.9 to 5.1 steps per second.

\section{Ablation on Distance Metric}
\label{ap:ablation:distance}
\begin{figure}[t]
    \centering
        \begin{subfigure}[t]{0.49\textwidth}
            \includegraphics[width=1\columnwidth,trim={0.3cm 3.3cm 0cm 0},clip]{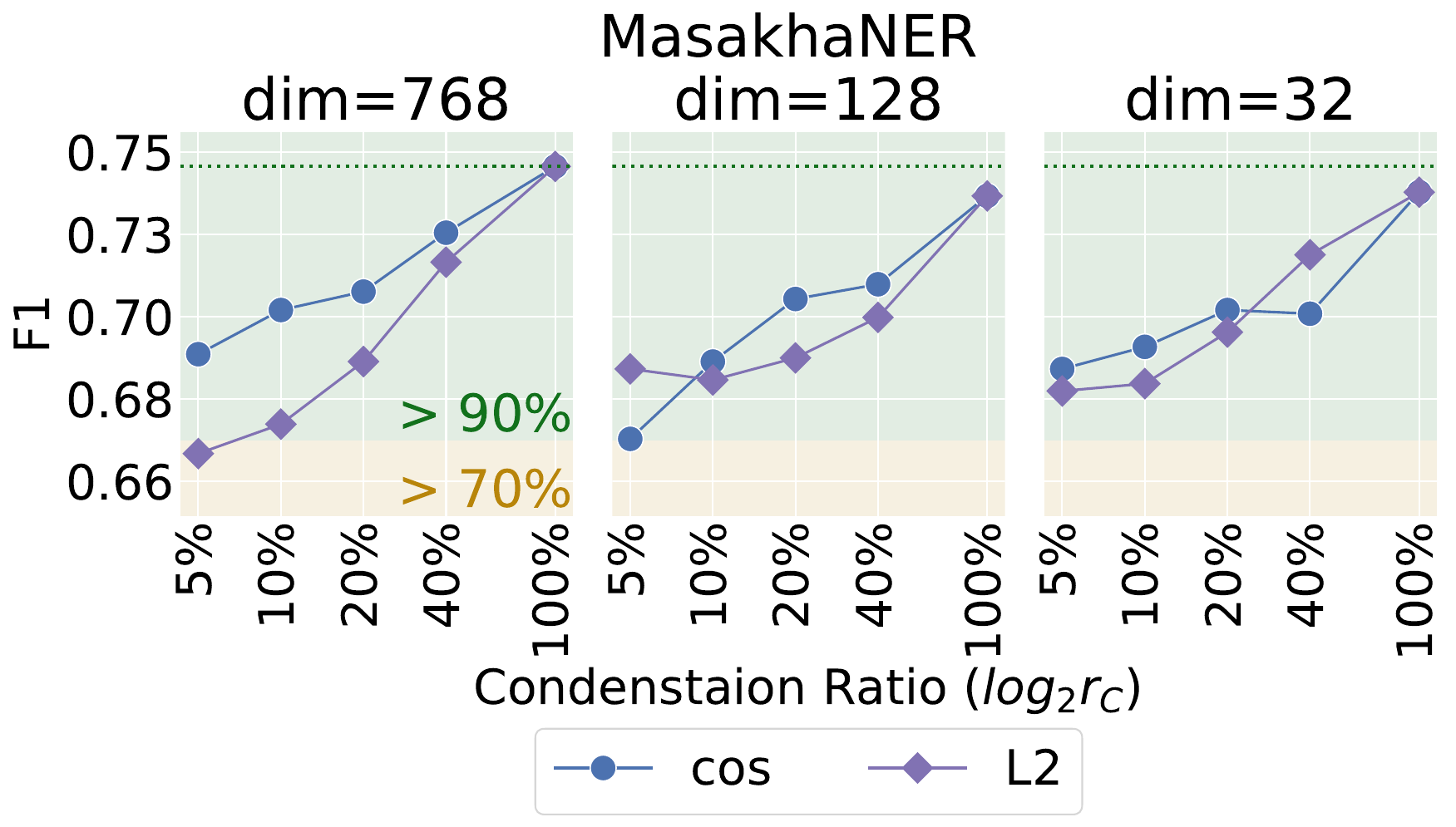}
        \end{subfigure}
        \begin{subfigure}[t]{0.49\textwidth}
            \includegraphics[width=1\columnwidth,trim={0.3cm 3.3cm 0cm 0},clip]{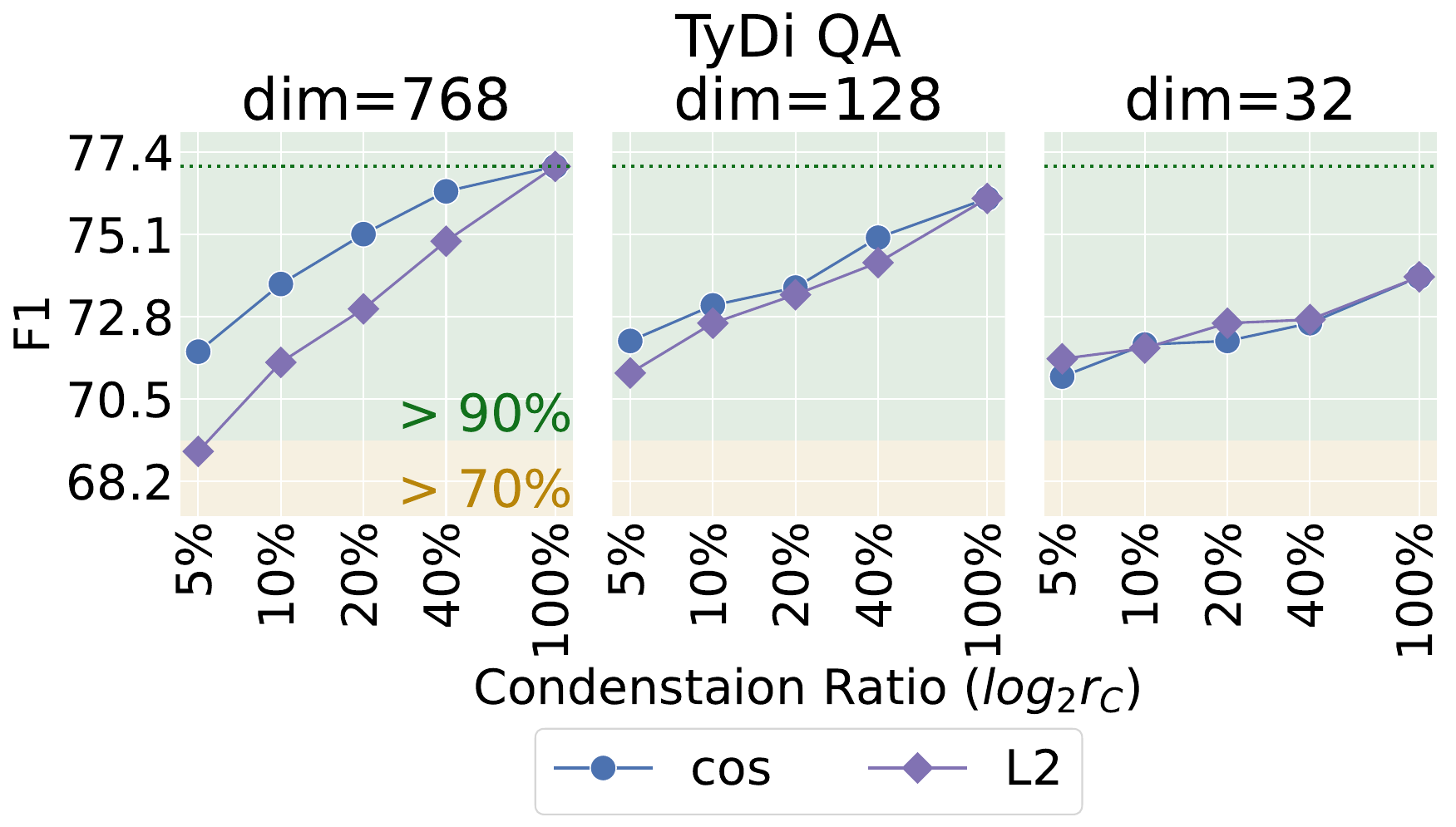}
        \end{subfigure}
        \begin{subfigure}[t]{0.49\textwidth}
            \includegraphics[width=1\columnwidth,trim={0.3cm 3.3cm 0cm 0},clip]{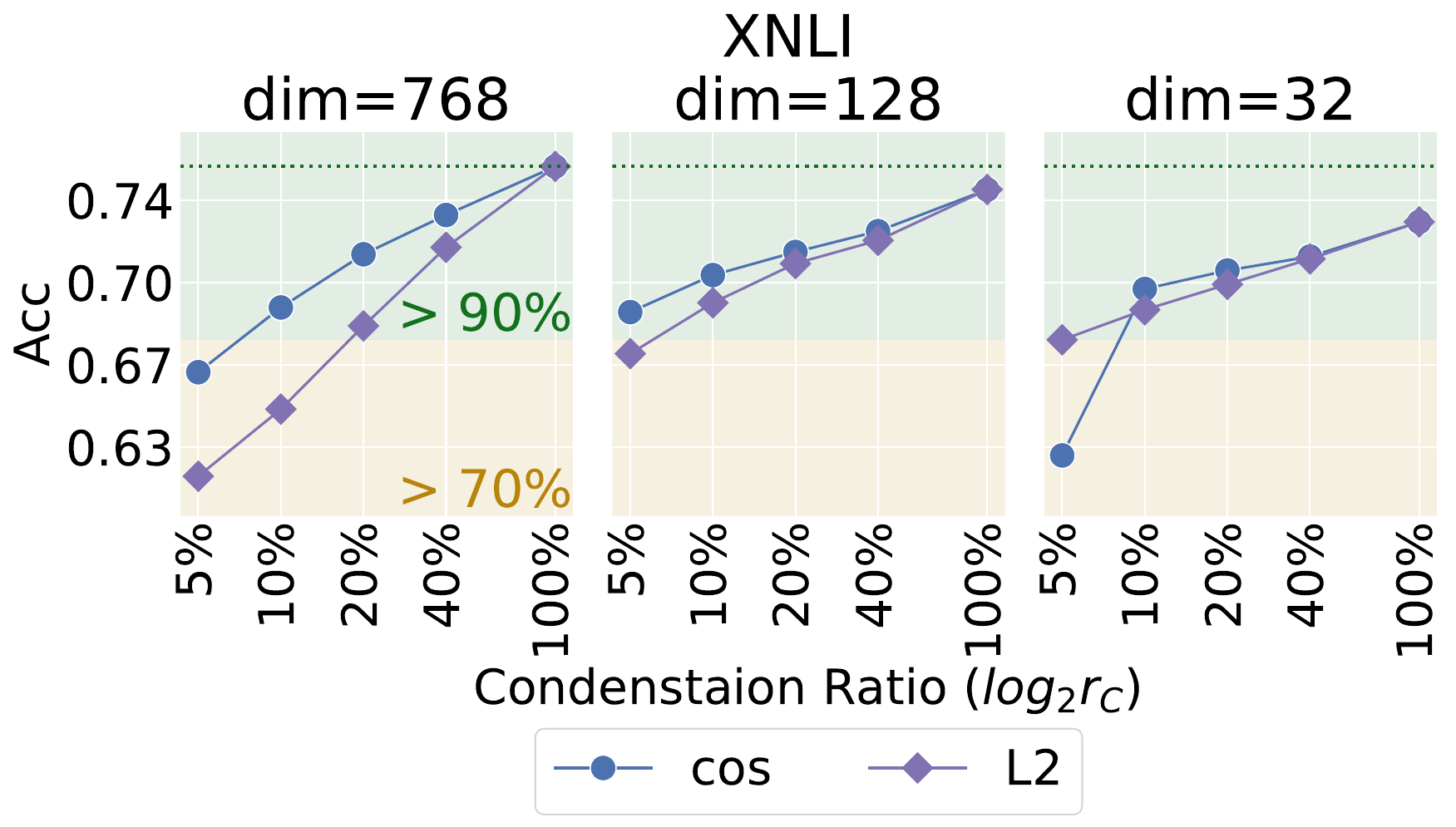}
        \end{subfigure}
        \begin{subfigure}[t]{0.49\textwidth}
                \includegraphics[width=1\columnwidth,trim={0.3cm 3.3cm 0cm 0},clip,right]{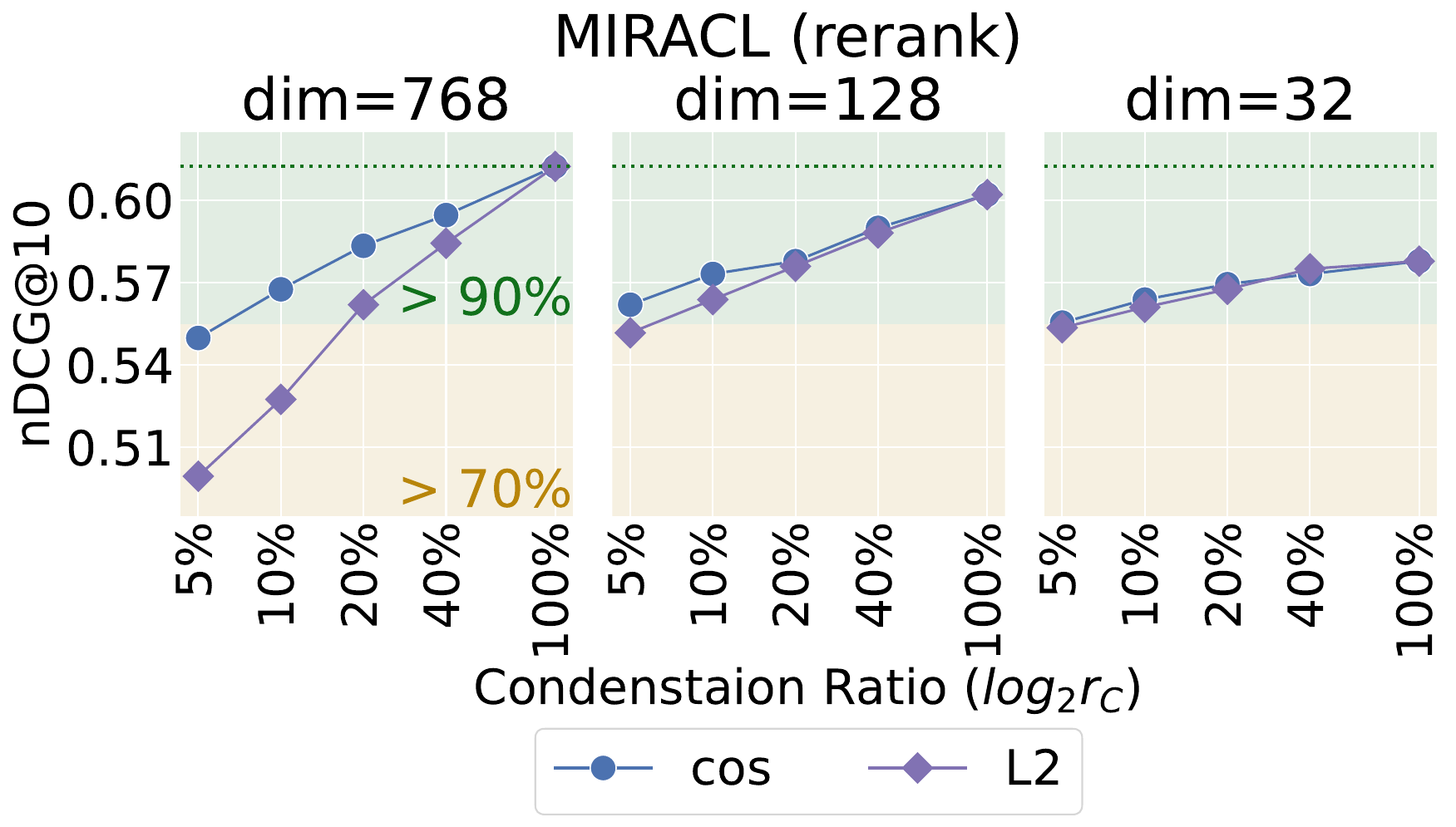}
        \end{subfigure}
        \begin{subfigure}[t]{0.49\textwidth}
            \includegraphics[width=1\columnwidth,trim={0.3cm 3.3cm 0cm 0},clip,left]{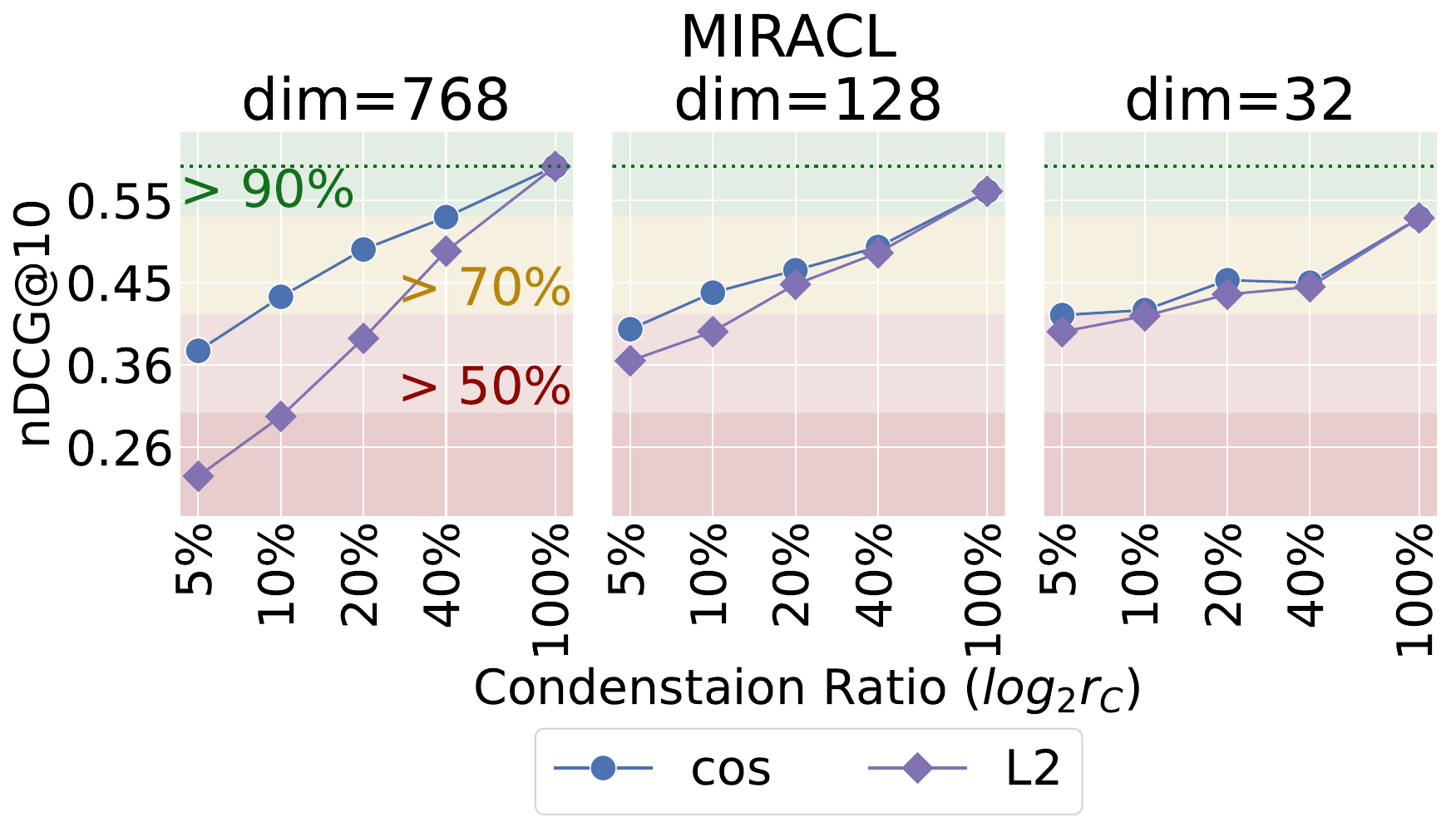}
        \end{subfigure}

        \begin{subfigure}[t]{0.49\textwidth}
            \includegraphics[width=0.95\columnwidth,trim={0cm 0cm 0cm 0},clip]{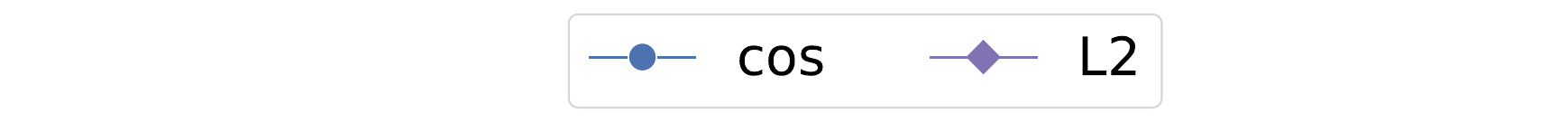}
        \end{subfigure}

    \caption{
        Comparison of L2 versus cosine distance when using \kmeans to group the subwords.
    }
    \label{fig:ablation:distance}
\end{figure}

We compare the results using cosine versus Euclidean distance in \autoref{fig:ablation:distance},
where the results show that grouping on Euclidean distance greatly underperformance cosine distance especially at higher dimension ($d=768$). 
We interpret this as that the vector norm is an undesired feature when pursuing the semantic similarity between the subwords, 
which amplifies the distance between semantic similar subword at high dimension.


\section{Grouping via Bilingual Lexicons}
\label{ap:sec:bilingual}
As an intuitive alternative to grouping via \kmeans,
we explored to group the subwords via the ground-truth bilingual lexicons in the preliminary experiments,
finding that it has limited coverage on the subwords, thus grouping ratios,
and also underperforming the \kmeans-based grouping.
See \autoref{tab:ap:bllex} for the results.

\begin{table}[t]
    \centering
    \begin{tabular}{lrr}
    \toprule
    {\bf Method} & $r_G$ & {\bf nDCG@10} \\
    \midrule
    Oracle & 100\% & 0.452 \\ 
    \midrule
    Grouping on MUSE (zh) & 99.2\%	& 0.357 \\
    Grouping on MUSE (5L) & 92.1\%	& 0.248	\\
    Grouping on MUSE (all) & 86.2\%	& 0.193	\\
    \midrule
    \kmeans & 40\% & 0.304 \\
    \bottomrule
    \end{tabular}

\caption{
    Results on \miracl (zh), comparing Grouping via Bilingual Lexicons vs \kmeans.
    Models are fine-tuned on MS MARCO, without CLSA or continual pretraining.
}
\label{tab:ap:bllex}
\end{table}

\section{Numerical Results of Experiments in Figure 2--6}
\label{ap:numerical-results}
\autoref{tab:ap:numerical-results} presents the numerical results of experiments in 
\autoref{fig:compareRandom768},
\ref{fig:768align},
\ref{fig:3dim},
\ref{fig:hidden-dim-vs-vocab},
and \ref{fig:backbones}.

\begin{table*}[]
    \resizebox{\linewidth}{!}{
    \begin{tabular}{rr|ll|rr|rr|rr|rr|rr}
    \toprule
    \multicolumn{1}{c}{$d$} & \multicolumn{1}{c}{$r_G$} & \textbf{w/ CLSA?} & \multicolumn{1}{c}{\textbf{w/ PT?}} & \multicolumn{2}{c}{\textbf{\masakhaner}} & \multicolumn{2}{c}{\textbf{\tydiqa}} & \multicolumn{2}{c}{\textbf{XNLI}} & \multicolumn{2}{c}{\textbf{MIRACL (cross-)}} & \multicolumn{2}{c}{\textbf{MIRACL (bi-)}} \\
    \multicolumn{1}{l}{} & \multicolumn{1}{l}{} &  &  & \multicolumn{1}{c}{\textbf{F1}} & \multicolumn{1}{c}{\textbf{perc}} & \multicolumn{1}{c}{\textbf{F1}} & \multicolumn{1}{c}{\textbf{perc}} & \multicolumn{1}{c}{\textbf{Acc}} & \multicolumn{1}{c}{\textbf{perc}} & \multicolumn{1}{l}{\textbf{nDCG@10}} & \multicolumn{1}{c}{\textbf{perc}} & \multicolumn{1}{c}{\textbf{nDCG@10}} & \multicolumn{1}{c}{\textbf{perc}} \\
    \midrule
    768 & 100\% & -- & \cross & 0.735 & \cellcolor[HTML]{E1EEE2}98.7\% & {\ul 76.9} & \cellcolor[HTML]{E1EEE2}99.9\% & 0.750 & \cellcolor[HTML]{E1EEE2}99.9\% & 0.613 & \cellcolor[HTML]{E1EEE2}99.5\% & 0.586 & \cellcolor[HTML]{E1EEE2}98.7\% \\
    768 & 100\% & -- & \tick & 0.745 & {\ul 100.0\%} & 77.0 & {\ul 100.0\%} & 0.751 & {\ul 100.0\%} & 0.616 & {\ul 100.0\%} & 0.594 & {\ul 100.0\%} \\
    \midrule
    \multicolumn{14}{c}{\large \it \autoref{fig:compareRandom768}} \\
    768 & 40\% & \cross & \cross & 0.704 & \cellcolor[HTML]{E1EEE2}94.5\% & 74.6 & \cellcolor[HTML]{E1EEE2}96.9\% & 0.725 & \cellcolor[HTML]{E1EEE2}96.5\% & 0.584 & \cellcolor[HTML]{E1EEE2}94.8\% & 0.493 & \cellcolor[HTML]{F7F0E0}83.0\% \\
    768 & 20\% & \cross & \cross & 0.698 & \cellcolor[HTML]{E1EEE2}93.7\% & {\ul 72.6} & \cellcolor[HTML]{E1EEE2}94.3\% & 0.702 & \cellcolor[HTML]{E1EEE2}93.5\% & 0.553 & \cellcolor[HTML]{F7F0E0}89.8\% & 0.436 & \cellcolor[HTML]{F7F0E0}73.4\% \\
    768 & 10\% & \cross & \cross & 0.680 & \cellcolor[HTML]{E1EEE2}91.3\% & 71.3 & \cellcolor[HTML]{E1EEE2}92.6\% & 0.678 & \cellcolor[HTML]{E1EEE2}90.3\% & 0.523 & \cellcolor[HTML]{F7F0E0}84.9\% & 0.372 & \cellcolor[HTML]{EFDFDF}62.6\% \\
    768 & 5\% & \cross & \cross & 0.659 & \cellcolor[HTML]{F7F0E0}88.5\% & 70.2 & \cellcolor[HTML]{E1EEE2}91.2\% & 0.650 & \cellcolor[HTML]{F7F0E0}86.6\% & 0.495 & \cellcolor[HTML]{F7F0E0}80.4\% & 0.306 & \cellcolor[HTML]{EFDFDF}51.5\% \\

    768 & 40\% & \cross & \tick & 0.727 & \cellcolor[HTML]{E1EEE2}97.6\% & {\ul 76.3} & \cellcolor[HTML]{E1EEE2}99.1\% & 0.730 & \cellcolor[HTML]{E1EEE2}97.2\% & 0.597 & \cellcolor[HTML]{E1EEE2}96.9\% & 0.533 & \cellcolor[HTML]{F7F0E0}89.7\% \\
    768 & 20\% & \cross & \tick & 0.711 & \cellcolor[HTML]{E1EEE2}95.4\% & 75.1 & \cellcolor[HTML]{E1EEE2}97.5\% & 0.713 & \cellcolor[HTML]{E1EEE2}94.9\% & 0.585 & \cellcolor[HTML]{E1EEE2}95.0\% & 0.494 & \cellcolor[HTML]{F7F0E0}83.2\% \\
    768 & 10\% & \cross & \tick & 0.706 & \cellcolor[HTML]{E1EEE2}94.8\% & 73.7 & \cellcolor[HTML]{E1EEE2}95.7\% & 0.690 & \cellcolor[HTML]{E1EEE2}91.9\% & 0.568 & \cellcolor[HTML]{E1EEE2}92.2\% & 0.437 & \cellcolor[HTML]{F7F0E0}73.6\% \\
    768 & 5\% & \cross & \tick & 0.694 & \cellcolor[HTML]{E1EEE2}93.2\% & {\ul 71.8} & \cellcolor[HTML]{E1EEE2}93.2\% & 0.662 & \cellcolor[HTML]{F7F0E0}88.1\% & 0.549 & \cellcolor[HTML]{F7F0E0}89.1\% & 0.372 & \cellcolor[HTML]{EFDFDF}62.6\% \\

    \midrule
    \multicolumn{14}{c}{\large \it \autoref{fig:768align}} \\
    768 & 40\% & \tick & \cross & 0.712 & \cellcolor[HTML]{E1EEE2}95.6\% & 74.7 & \cellcolor[HTML]{E1EEE2}97.0\% & 0.729 & \cellcolor[HTML]{E1EEE2}97.1\% & 0.580 & \cellcolor[HTML]{E1EEE2}94.2\% & 0.487 & \cellcolor[HTML]{F7F0E0}82.0\% \\
    768 & 20\% & \tick & \cross & 0.705 & \cellcolor[HTML]{E1EEE2}94.6\% & 73.0 & \cellcolor[HTML]{E1EEE2}94.8\% & 0.715 & \cellcolor[HTML]{E1EEE2}95.2\% & 0.563 & \cellcolor[HTML]{E1EEE2}91.4\% & 0.443 & \cellcolor[HTML]{F7F0E0}74.6\% \\
    768 & 10\% & \tick & \cross & 0.686 & \cellcolor[HTML]{E1EEE2}92.1\% & {\ul 71.3} & \cellcolor[HTML]{E1EEE2}92.6\% & 0.697 & \cellcolor[HTML]{E1EEE2}92.8\% & 0.542 & \cellcolor[HTML]{F7F0E0}88.0\% & 0.389 & \cellcolor[HTML]{EFDFDF}65.5\% \\
    768 & 5\% & \tick & \cross & 0.679 & \cellcolor[HTML]{E1EEE2}91.1\% & 70.7 & \cellcolor[HTML]{E1EEE2}91.8\% & 0.677 & \cellcolor[HTML]{E1EEE2}90.1\% & 0.516 & \cellcolor[HTML]{F7F0E0}83.8\% & 0.330 & \cellcolor[HTML]{EFDFDF}55.6\% \\

    768 & 40\% & \tick & \tick & 0.724 & \cellcolor[HTML]{E1EEE2}97.2\% & {\ul 76.5} & \cellcolor[HTML]{E1EEE2}99.4\% & 0.741 & \cellcolor[HTML]{E1EEE2}98.7\% & 0.602 & \cellcolor[HTML]{E1EEE2}97.7\% & 0.540 & \cellcolor[HTML]{E1EEE2}90.9\% \\
    768 & 20\% & \tick & \tick & 0.722 & \cellcolor[HTML]{E1EEE2}96.9\% & 74.8 & \cellcolor[HTML]{E1EEE2}97.1\% & 0.726 & \cellcolor[HTML]{E1EEE2}96.7\% & 0.590 & \cellcolor[HTML]{E1EEE2}95.8\% & 0.506 & \cellcolor[HTML]{F7F0E0}85.2\% \\
    768 & 10\% & \tick & \tick & 0.714 & \cellcolor[HTML]{E1EEE2}95.8\% & 74.1 & \cellcolor[HTML]{E1EEE2}96.2\% & 0.707 & \cellcolor[HTML]{E1EEE2}94.1\% & 0.570 & \cellcolor[HTML]{E1EEE2}92.5\% & 0.442 & \cellcolor[HTML]{F7F0E0}74.4\% \\
    768 & 5\% & \tick & \tick & 0.691 & \cellcolor[HTML]{E1EEE2}92.8\% & {\ul 73.0} & \cellcolor[HTML]{E1EEE2}94.8\% & 0.690 & \cellcolor[HTML]{E1EEE2}91.9\% & 0.566 & \cellcolor[HTML]{E1EEE2}91.9\% & 0.398 & \cellcolor[HTML]{EFDFDF}67.0\% \\

    \midrule
    \multicolumn{14}{c}{\large \it \autoref{fig:3dim}} \\
    128 & 40\% & \tick & \tick & 0.717 & \cellcolor[HTML]{E1EEE2}96.2\% & {\ul 75.2} & \cellcolor[HTML]{E1EEE2}97.7\% & 0.725 & \cellcolor[HTML]{E1EEE2}96.5\% & 0.591 & \cellcolor[HTML]{E1EEE2}95.9\% & 0.509 & \cellcolor[HTML]{F7F0E0}85.7\% \\
    128 & 20\% & \tick & \tick & 0.714 & \cellcolor[HTML]{E1EEE2}95.8\% & 74.5 & \cellcolor[HTML]{E1EEE2}96.8\% & 0.721 & \cellcolor[HTML]{E1EEE2}96.0\% & 0.585 & \cellcolor[HTML]{E1EEE2}95.0\% & 0.469 & \cellcolor[HTML]{F7F0E0}79.0\% \\
    128 & 10\% & \tick & \tick & 0.698 & \cellcolor[HTML]{E1EEE2}93.7\% & 73.3 & \cellcolor[HTML]{E1EEE2}95.2\% & 0.708 & \cellcolor[HTML]{E1EEE2}94.3\% & 0.577 & \cellcolor[HTML]{E1EEE2}93.7\% & 0.451 & \cellcolor[HTML]{F7F0E0}75.9\% \\
    128 & 5\% & \tick & \tick & 0.695 & \cellcolor[HTML]{E1EEE2}93.3\% & {\ul 72.9} & \cellcolor[HTML]{E1EEE2}94.7\% & 0.694 & \cellcolor[HTML]{E1EEE2}92.4\% & 0.568 & \cellcolor[HTML]{E1EEE2}92.2\% & 0.415 & \cellcolor[HTML]{EFDFDF}69.9\% \\
    32 & 40\% & \tick & \tick & 0.712 & \cellcolor[HTML]{E1EEE2}95.6\% & 74.7 & \cellcolor[HTML]{E1EEE2}97.0\% & 0.729 & \cellcolor[HTML]{E1EEE2}97.1\% & 0.580 & \cellcolor[HTML]{E1EEE2}94.2\% & 0.487 & \cellcolor[HTML]{F7F0E0}82.0\% \\
    32 & 20\% & \tick & \tick & 0.705 & \cellcolor[HTML]{E1EEE2}94.6\% & {\ul 73.0} & \cellcolor[HTML]{E1EEE2}94.8\% & 0.715 & \cellcolor[HTML]{E1EEE2}95.2\% & 0.563 & \cellcolor[HTML]{E1EEE2}91.4\% & 0.443 & \cellcolor[HTML]{F7F0E0}74.6\% \\
    32 & 10\% & \tick & \tick & 0.686 & \cellcolor[HTML]{E1EEE2}92.1\% & 71.3 & \cellcolor[HTML]{E1EEE2}92.6\% & 0.697 & \cellcolor[HTML]{E1EEE2}92.8\% & 0.542 & \cellcolor[HTML]{F7F0E0}88.0\% & 0.389 & \cellcolor[HTML]{EFDFDF}65.5\% \\
    32 & 5\% & \tick & \tick & 0.679 & \cellcolor[HTML]{E1EEE2}91.1\% & 70.7 & \cellcolor[HTML]{E1EEE2}91.8\% & 0.677 & \cellcolor[HTML]{E1EEE2}90.1\% & 0.516 & \cellcolor[HTML]{F7F0E0}83.8\% & 0.330 & \cellcolor[HTML]{EFDFDF}55.6\% \\

    \midrule
    \multicolumn{14}{c}{\large \it \autoref{fig:hidden-dim-vs-vocab}} \\
    128 & 100\% & -- & \tick & 0.737 & \cellcolor[HTML]{E1EEE2}98.9\% & 76.1 & \cellcolor[HTML]{E1EEE2}98.8\% & 0.741 & \cellcolor[HTML]{E1EEE2}98.7\% & 0.605 & \cellcolor[HTML]{E1EEE2}98.2\% & 0.564 & \cellcolor[HTML]{E1EEE2}94.9\% \\
    32 & 100\% & -- & \tick & 0.735 & \cellcolor[HTML]{E1EEE2}98.7\% & {\ul 76.9} & \cellcolor[HTML]{E1EEE2}99.9\% & 0.750 & \cellcolor[HTML]{E1EEE2}99.9\% & 0.613 & \cellcolor[HTML]{E1EEE2}99.5\% & 0.586 & \cellcolor[HTML]{E1EEE2}98.7\% \\
    8 & 100\% & -- & \tick & 0.639 & \cellcolor[HTML]{F7F0E0}85.8\% & 47.8 & \cellcolor[HTML]{EFDFDF}62.1\% & 0.626 & \cellcolor[HTML]{F7F0E0}83.4\% & 0.414 & \cellcolor[HTML]{EFDFDF}67.2\% & 0.276 & \cellcolor[HTML]{F6CCCC}46.5\% \\
    2 & 100\% & -- & \tick & 0.374 & \cellcolor[HTML]{EFDFDF}50.2\% & 29.4 & \cellcolor[HTML]{F6CCCC}38.2\% & 0.333 & \cellcolor[HTML]{F6CCCC}44.3\% & 0.123 & \cellcolor[HTML]{F6CCCC}20.0\% & 0.064 & \cellcolor[HTML]{F6CCCC}10.8\% \\

    \midrule
    \multicolumn{14}{c}{\large \it \autoref{fig:backbones}: XLM-R base} \\
    768 & 100\% & \tick & \cross & 0.814 & {\ul 100.0\%} & {\ul 78.2} & {\ul 100.0\%} & 0.779 & {\ul 100.0\%} & 0.611 & {\ul 100.0\%} & 0.558 & {\ul 100.0\%} \\
    768 & 40\% & \tick & \cross & 0.783 & \cellcolor[HTML]{E1EEE2}96.2\% & 75.5 & \cellcolor[HTML]{E1EEE2}96.5\% & 0.753 & \cellcolor[HTML]{E1EEE2}96.7\% & 0.584 & \cellcolor[HTML]{E1EEE2}95.6\% & 0.484 & \cellcolor[HTML]{F7F0E0}86.7\% \\
    768 & 20\% & \tick & \cross & 0.739 & \cellcolor[HTML]{E1EEE2}90.8\% & 73.1 & \cellcolor[HTML]{E1EEE2}93.5\% & 0.735 & \cellcolor[HTML]{E1EEE2}94.4\% & 0.556 & \cellcolor[HTML]{E1EEE2}91.0\% & 0.431 & \cellcolor[HTML]{F7F0E0}77.2\% \\
    768 & 10\% & \tick & \cross & 0.717 & \cellcolor[HTML]{F7F0E0}88.1\% & {\ul 70.8} & \cellcolor[HTML]{E1EEE2}90.5\% & 0.701 & \cellcolor[HTML]{E1EEE2}90.0\% & 0.516 & \cellcolor[HTML]{F7F0E0}84.5\% & 0.374 & \cellcolor[HTML]{EFDFDF}67.0\% \\
    768 & 5\% & \tick & \cross & 0.695 & \cellcolor[HTML]{F7F0E0}85.4\% & 66.8 & \cellcolor[HTML]{F7F0E0}85.4\% & 0.669 & \cellcolor[HTML]{F7F0E0}85.9\% & 0.490 & 80.2\% & 0.314 & \cellcolor[HTML]{EFDFDF}56.3\% \\
    \midrule
    \multicolumn{14}{c}{\large \it \autoref{fig:backbones}: XLM-R large} \\
    1024 & 100\% & \tick & \cross & 0.792 & {\ul 100.0\%} & 80.4 & {\ul 100.0\%} & 0.844 & {\ul 100.0\%} & 0.645 & {\ul 100.0\%} & 0.598 & {\ul 100.0\%} \\
    1024 & 40\% & \tick & \cross & 0.792 & \cellcolor[HTML]{E1EEE2}100.0\% & 79.6 & \cellcolor[HTML]{E1EEE2}99.0\% & 0.809 & \cellcolor[HTML]{E1EEE2}95.9\% & 0.624 & \cellcolor[HTML]{E1EEE2}96.7\% & 0.547 & \cellcolor[HTML]{E1EEE2}91.5\% \\
    1024 & 20\% & \tick & \cross & 0.755 & \cellcolor[HTML]{E1EEE2}95.3\% & {\ul 78.3} & \cellcolor[HTML]{E1EEE2}97.4\% & 0.762 & \cellcolor[HTML]{E1EEE2}90.3\% & 0.595 & \cellcolor[HTML]{E1EEE2}92.2\% & 0.442 & \cellcolor[HTML]{F7F0E0}73.9\% \\
    1024 & 10\% & \tick & \cross & 0.715 & \cellcolor[HTML]{E1EEE2}90.3\% & 75.3 & \cellcolor[HTML]{E1EEE2}93.7\% & 0.716 & \cellcolor[HTML]{F7F0E0}84.8\% & 0.557 & \cellcolor[HTML]{F7F0E0}86.4\% & 0.384 & \cellcolor[HTML]{EFDFDF}64.2\% \\
    1024 & 5\% & \tick & \cross & 0.722 & \cellcolor[HTML]{E1EEE2}91.2\% & 71.7 & \cellcolor[HTML]{F7F0E0}89.2\% & 0.668 & \cellcolor[HTML]{F7F0E0}79.1\% & 0.531 & \cellcolor[HTML]{F7F0E0}82.3\% & 0.304 & \cellcolor[HTML]{EFDFDF}50.8\% \\
    \midrule
    \multicolumn{14}{c}{\large \it \autoref{fig:backbones}: XLM-V base} \\
    768 & 100\% & \tick & \cross & 0.828 & {\ul 100.0\%} & 79.2 & {\ul 100.0\%} & 0.792 & {\ul 100.0\%} & 0.621 & {\ul 100.0\%} & 0.600 & {\ul 100.0\%} \\
    768 & 40\% & \tick & \cross & 0.773 & \cellcolor[HTML]{E1EEE2}93.4\% & {\ul 75.2} & \cellcolor[HTML]{E1EEE2}94.9\% & 0.755 & \cellcolor[HTML]{E1EEE2}95.3\% & 0.582 & \cellcolor[HTML]{E1EEE2}93.7\% & 0.463 & \cellcolor[HTML]{F7F0E0}77.2\% \\
    768 & 20\% & \tick & \cross & 0.719 & \cellcolor[HTML]{F7F0E0}86.8\% & 73.8 & \cellcolor[HTML]{E1EEE2}93.2\% & 0.722 & \cellcolor[HTML]{E1EEE2}91.2\% & 0.553 & \cellcolor[HTML]{F7F0E0}89.0\% & 0.398 & \cellcolor[HTML]{EFDFDF}66.3\% \\
    768 & 10\% & \tick & \cross & 0.707 & \cellcolor[HTML]{F7F0E0}85.4\% & 71.7 & \cellcolor[HTML]{E1EEE2}90.5\% & 0.690 & \cellcolor[HTML]{F7F0E0}87.1\% & 0.522 & \cellcolor[HTML]{F7F0E0}84.1\% & 0.327 & \cellcolor[HTML]{EFDFDF}54.5\% \\
    768 & 5\% & \tick & \cross & 0.676 & \cellcolor[HTML]{F7F0E0}81.6\% & 68.8 & \cellcolor[HTML]{F7F0E0}86.9\% & 0.656 & \cellcolor[HTML]{F7F0E0}82.8\% & 0.497 & \cellcolor[HTML]{F7F0E0}80.0\% & 0.277 & \cellcolor[HTML]{F6CCCC}46.2\% \\
    \bottomrule
    \end{tabular}
}
\caption{
    Numerical results of experiments in all figures, where each number is an averaged result of all languages per benchmark. 
    We skip the per-language score due to the space limit.
    $d$:\ word embedding dimension;
    $r_G$:\ grouping ratio;
    {\bf ``cross-''}: cross-encoder
    {\bf ``bi-''}: bi-encoder
    \textbf{perc}:\ the relative performance to the oracle results (i.e., the second row for mBERT, the corresponding 100\% rows for the other backbones) 
    The background colors 
    indicate the relative performance to the oracle results:\ \greencolor: >90\%, \yellowcolor: 70\%--90\%, \redcolor: 50\%--70\% and the dark red color means <50\%. 
    Better viewed in colors.
}
\label{tab:ap:numerical-results}
\end{table*}

\section{More Inspection Examples}
\label{ap:inspection}
\begin{figure*}[b]
    \includegraphics[width=\textwidth]{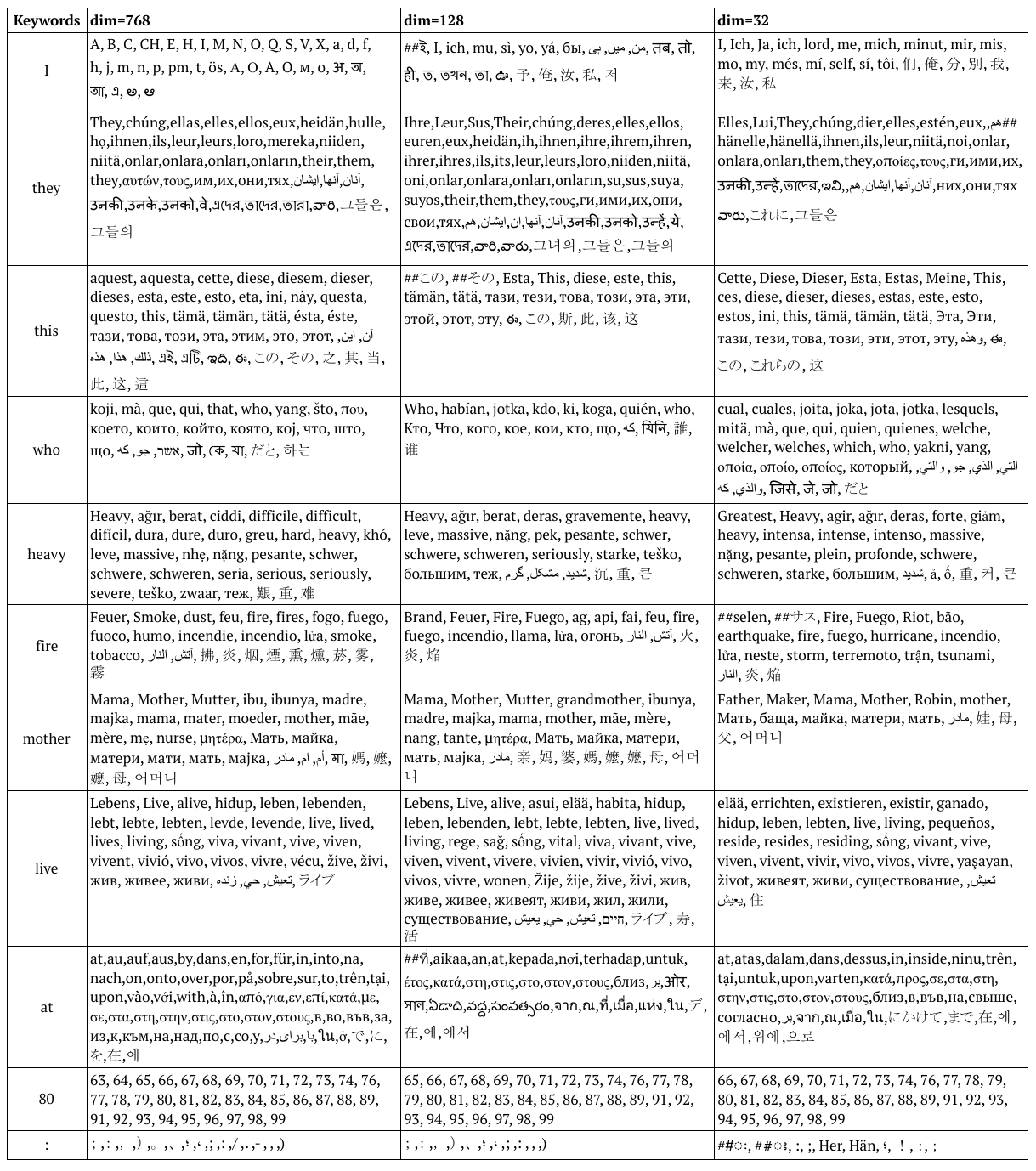} 
    \caption{More examples of the grouped subwords on mBERT with CLSA}
    \label{fig:ap:inspection}
\end{figure*}

\autoref{fig:ap:inspection} shows more examples of the grouped subwords additional to \autoref{fig:inspectionCluster}.

\end{document}